%% file: paper.tex
\documentclass[10pt,twocolumn,letterpaper]{article}

\usepackage[pagenumbers]{cvpr} 
\usepackage[accsupp]{axessibility} 
\usepackage{amsfonts}
\usepackage{amsthm}
\usepackage{amsmath}
\usepackage{amssymb}
\usepackage{bm}
\usepackage{xspace}
\usepackage{nicefrac}
\usepackage{cases}
\usepackage{booktabs}
\usepackage{threeparttable}
\usepackage{enumitem}
\usepackage{colortbl}
\usepackage{multirow}
\usepackage{stfloats}
\usepackage{xcolor}
\usepackage{color}

\usepackage{graphicx}
\usepackage{pifont}
\usepackage{makecell}
\usepackage{wrapfig}
\usepackage{draftfigure}
\usepackage{lipsum}

\usepackage[pagebackref,breaklinks,colorlinks]{hyperref}

\usepackage{algorithm}
\usepackage{algpseudocode}

\algrenewcommand{\algorithmiccomment}[1]{\quad\# #1}

\newcommand{\Blue}[1]{{\color{blue}#1}}

\newcommand{\cmark}{\ding{51}}%
\newcommand{\xmark}{\ding{55}}%

\theoremstyle{definition}

\usepackage[capitalize]{cleveref}
\crefname{section}{Sec.}{Secs.}
\Crefname{section}{Section}{Sections}
\Crefname{table}{Table}{Tables}
\crefname{table}{Tab.}{Tabs.}

\def\cvprPaperID{1458} 
\def\confName{CVPR}
\def\confYear{2022}

\begin{document}

\title{Bootstrapping ViTs: Towards Liberating Vision Transformers from Pre-training}
\makeatletter
\newcommand{\myfnsymbol}[1]{%
  \expandafter\@myfnsymbol\csname c@#1\endcsname
}
\newcommand{\@myfnsymbol}[1]{%
  \ifcase #1
  \or \TextOrMath{\textasteriskcentered}{*}
  \or \TextOrMath{\textdagger}{\dagger}
  \fi
}
\makeatother
\author {%
    Haofei Zhang\textsuperscript{\rm{1},$*$},
    Jiarui Duan\textsuperscript{\rm{1},$*$},
    Mengqi Xue\textsuperscript{\rm{1}},
    Jie Song\textsuperscript{\rm{1},$\dagger$},
    Li Sun\textsuperscript{\rm{1}},
    Mingli Song\textsuperscript{\rm{1},\rm{2}} \\
    \textsuperscript{\rm{1}}Zhejiang University \quad \textsuperscript{\rm{2}} Shanghai Institute for Advanced Study of Zhejiang University
}

\renewcommand{\thefootnote}{\myfnsymbol{footnote}}
\maketitle
\footnotetext[1]{Equal contribution}
\footnotetext[2]{Corresponding author, email: sjie@zju.edu.cn}%

\setcounter{footnote}{0}
\renewcommand{\thefootnote}{\arabic{footnote}}

\begin{abstract}
\input{main/abstract.tex}
\end{abstract}

\section{Introduction}
\input{main/introduction.tex}

\section{Related Work}
\input{main/related_work.tex}

\section{Method}
\input{main/method.tex}

\section{Experiments}
\input{main/experiments.tex}

\section{Conclusion and Future Work}
\input{main/conclusion.tex}

\paragraph{Acknowledgements.}
This work is supported by Key Research and Development Program of Zhejiang Province (2020C01024), CCF-Baidu Open Fund (NO.2021PP15002000), National Natural Science Foundation of China (62106220, U20B2066), Ningbo Natural Science Foundation (2021J189), the Starry Night Science Fund of Zhejiang University Shanghai Institute for Advanced Study (Grant No. SN-ZJU-SIAS-001), and the Fundamental Research Funds for the Central Universities.

\clearpage
{\small
\bibliographystyle{ieee_fullname}
\bibliography{references/cnn.bib,references/transformers.bib,references/traditional.bib}
}
\clearpage
\appendix
\section{Convolution in Matrix Form}
\label{appendix:conv}
\input{appendices/conv.tex}

\section{Intermediate Supervision Analysis}
\label{appendix:intermediate}
\input{appendices/interdimate_sup.tex}

\section{Visualization of MHSA}
\label{appendix:vis_mhsa}
\input{appendices/vis_mhsa.tex}

\section{Visualization of Intermediate Feature}
\label{appendix:vis_seq}
\input{appendices/vis_seq.tex}

\section{Network Architecture of Agent CNNs}
\label{appendix:arch}
\input{appendices/arch.tex}
\clearpage
\input{appendices/visualization.tex}

\end{document}

%% file: main/abstract.tex
Recently, vision Transformers~(ViTs) are developing rapidly and starting to challenge the domination of convolutional neural networks~(CNNs) in the realm of computer vision~(CV).
With the general-purpose Transformer architecture replacing the hard-coded inductive biases of convolution, ViTs have surpassed CNNs, especially in data-sufficient circumstances.
However, ViTs are prone to over-fit on small datasets and thus rely on large-scale pre-training, which expends enormous time.
In this paper, we strive to liberate ViTs from pre-training by introducing CNNs' inductive biases back to ViTs while preserving their network architectures for higher upper bound and setting up more suitable optimization objectives.
To begin with, an agent CNN is designed based on the given ViT with inductive biases.
Then a bootstrapping training algorithm is proposed to jointly optimize the agent and ViT with weight sharing, during which the ViT learns inductive biases from the intermediate features of the agent.
Extensive experiments on CIFAR-10/100 and ImageNet-1k with limited training data have shown encouraging results that the inductive biases help ViTs converge significantly faster and outperform conventional CNNs with even fewer parameters.
Our code is publicly available at \url{https://github.com/zhfeing/Bootstrapping-ViTs-pytorch}.

%% file: main/introduction.tex
The great successes of the convolutional neural networks~(CNNs)~\cite{krizhevsky2012-imagenet,szegedy2015-going,he2016-deep,huang2017-densely} have liberated researchers from handcrafting visual features~\cite{lowe2004-distinctive-sift,dalal2005-histograms}.
By means of the inductive biases~\cite{cohen2016-inductive}, \ie, focusing on the localized features and weight sharing, CNNs are potent tools for tackling visual recognition tasks~\cite{he2016-deep,ren2015-faster,chen2017-rethinking}.
Nevertheless, such biases have constrained their abilities towards building deeper and larger models, as they have ignored the long-range dependencies~\cite{dosovitskiy2021-vit,d2021-convit}.

In recent years, \textit{Transformers}~\cite{NIPS2017-attention} have been proposed for replacing inductive biases with a general-purpose network architecture in natural language processing~(NLP).
Exclusively relying on multi-head attention mechanisms~(MHA), Transformers have the inborn capability to capture the global dependencies within language tokens and have become the \textit{de facto} preferred data-driven models in NLP~\cite{brown2020-gpt,devlin2018-bert,radford2019-language}.
Inspired by this, a growing number of researchers have introduced the Transformer architecture into the realm of computer vision~(CV)~\cite{carion2020-detr,dosovitskiy2021-vit,touvron2020-deit,zheng2021-rethinking}.
It turns out an encouraging discovery that vision Transformers~(ViTs) outperform state-of-the-art~(SOTA) CNNs by a large margin with a similar amount of parameters.

\input{floats/figure_intro.tex}

Despite the appealing achievements, ViTs suffer from poor performance, especially without adequate annotations or strong data augmentation strategies~\cite{dosovitskiy2021-vit,touvron2020-deit, chen2021-vit-outperform}. The reasons for this circumstance are two fold:
on the one hand, the widely adopted multi-head self-attention mechanisms~(MHSA) in ViTs have dense connections against convolution~\cite{cordonnier2019-relationship}, which is hard to optimize without prior knowledge;
on the other hand, Chen~\etal~\cite{chen2021-vit-outperform} have illustrated that ViTs tend to converge to minima with sharp regions, usually related to limited generalization capability and overfitting problems~\cite{keskar2016-large,chen2020-stabilizing}.
Therefore, the typical training scheme of Transformers in NLP~\cite{devlin2018-bert,brown2020-gpt} relies on the large-scale pre-training and then fine-tuning for downstream tasks, which consume enormous GPU~(TPU) time and energy~\cite{dosovitskiy2021-vit,touvron2020-deit,carion2020-detr}.
For instance, Dosovitskiy~\etal~\cite{dosovitskiy2021-vit} spend thousands of TPU days to pre-train a ViT with 303M images. Spontaneously, it raises the following question: how can we optimize ViTs efficiently without pre-training.

To the best of our knowledge, existing approaches focused on the problem can be mainly divided into two parts. The first line of approaches attempts to bring inductive biases back into Transformers, such as sparse attention~\cite{correia2019-adaptively,chen2021-chasing,kim2021-rethinking} and token aggregation~\cite{yuan2021-tokenvit}.
Such heuristic modifications to ViTs will inevitably lead to the sophisticated tuning of plenty of hyperparameters.
The second line of approaches~\cite{chen2021-vit-outperform,touvron2020-deit,jiao2019-tinybert} aims at constructing suitable training schemes for Transformers, which helps them converge with better generalization ability.
In particular, Chen~\etal~\cite{chen2021-vit-outperform} utilize the sharpness-aware minimizer~(SAM)~\cite{foret2020-sharpness} to find smooth minima, while~\cite{touvron2020-deit,jiao2019-tinybert} optimize a Transformer by distilling knowledge from a pre-trained teacher.
Nonetheless, these methods still require pre-training on mid-sized datasets, such as ImageNet-1k~\cite{krizhevsky2012-imagenet}, and how to efficiently train ViTs with relatively small datasets from scratch remains an open question.

Motivated by the distillation approaches~\cite{hinton2015-distilling,touvron2020-deit,jiao2019-tinybert} that utilize a teacher model to guide the optimization direction of the student, in this paper, we strive to make one step further towards optimizing ViTs with the help of an agent CNN, which also learns from scratch along with the ViT.
Our goal is to inject the inductive biases from the agent CNN into the ViT without modifying its architecture and design a more friendly optimization process so that ViTs can be customized on small-scale datasets without pre-training.

To this end, we propose a novel optimization strategy for training vision Transformers in the \textit{bootstrapping} form so that even without pre-training on mid-sized datasets or strong data augmentations, ViTs can still be competitive when lack of training data.
Specifically, as shown in~\cref{fig:intro}, we first propose an \textit{agent} CNN designed corresponding to the given ViT, and with the inductive biases, the agent will converge faster than the ViT.
Then we jointly optimize the ViT along with the agent in the mutual learning framework~\cite{zhang2018-mutual}, where the intermediate features of the agent supervise the ViT with the inductive biases for fast convergence.
In order to reduce the training burden, we further share the parameters of the ViT to the agent and propose a bootstrapping learning algorithm to update the shared parameters.
We have conducted extensive experiments on CIFAR-10/100 datasets~\cite{krizhevsky2009-cifar} and ImageNet-1k~\cite{krizhevsky2012-imagenet} under the lack-of-data settings.
Experimental results demonstrate that:
(1) our method has successfully injected the inductive biases to ViTs as they converge significantly faster than training from scratch and eventually surpass both the agents and SOTA CNNs;
(2) the bootstrapping learning method can efficiently optimize the shared weights without the extra set of parameters.

Our contributions are summarized as three folds:
\begin{enumerate}
    \item We propose agent CNNs constructed based on standard ViTs, for training ViTs efficiently with shared weights and inductive biases.
    \item We propose a novel bootstrapping optimization algorithm to optimize the shared parameters.
    \item Our experiments show that ViTs can outperform the SOTA CNNs even without pre-training by adopting both inductive biases and suitable optimizing goals.
\end{enumerate}

%% file: floats/figure_intro.tex
\begin{figure}[!t]
\centering%
    \includegraphics[width=\linewidth]{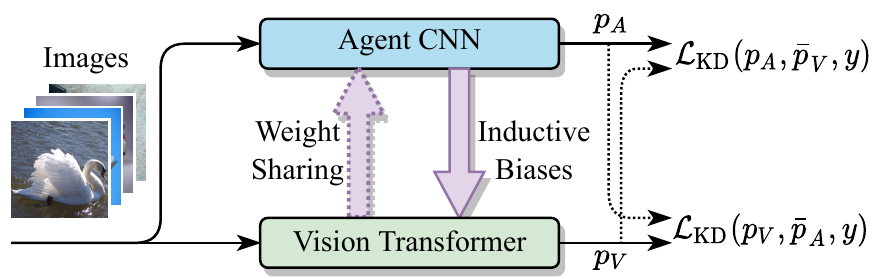}
\caption{Illustration of our proposed method for optimizing vision Transformers efficiently without pre-training. An agent CNN is constructed according to the network architecture of the ViT with shared weights, and the ViT learns inductive biases from intermediate features and predictions of the agent.}%
\label{fig:intro}%
\end{figure}

%% file: main/related_work.tex
\subsection{Vision Transformers}
With the powerful self-attention mechanisms, the Transformer~\cite{NIPS2017-attention} has been the SOTA and preferred model in NLP~\cite{devlin2018-bert, brown2020-gpt, brown2020-gpt3}.
Inspired by the impressive success of Transformers in NLP, researchers are starting to introduce Transformers for tackling CV tasks.
ViT~\cite{dosovitskiy2021-vit} is a groundbreaking work that utilizes the pure Transformer architecture for image classification and has achieved great successes.
The variants of the ViT~\cite{touvron2020-deit, liu2021-swin, yuan2021-tokenvit, wu2021-cvt, chen2021-visformer, zhou2021deepvit} are further utilized for more complex CV tasks, \eg, semantic segmentation~\cite{strudel2021segmenter, zheng2021-rethinking} and object detection~\cite{carion2020-detr, dai2021-updetr}.
However, ViTs rely on large-scale pre-training and have shown poor performance with limited training data.
To address this problem, some approaches try to introduce inductive biases to ViTs with heuristic modification, \eg, sparse attention~\cite{correia2019-adaptively,chen2021-chasing,kim2021-rethinking}, token aggregation~\cite{yuan2021-tokenvit}. The others aim to propose novel training schemes tailored for Transformers~\cite{chen2021-vit-outperform,touvron2020-deit,jiao2019-tinybert}.
Nonetheless, these methods still require pre-training on mid-scale datasets, such as ImageNet-1k.
It still remains an open question of how to optimize ViTs efficiently without pre-training, especially on small-scale datasets.
To resolve this problem, we strive to inject CNNs' inductive biases into ViTs without modifications to the network architecture.

\subsection{Knowledge Distillation}
\textit{Knowledge distillation}~(KD)~\cite{hinton2015-distilling} is an effective model compression technique which the hidden knowledge of the teacher is transferred to the student by supervising with soft labels.
To sufficiently transfer the knowledge, FitNets~\cite{romero2014fitnets} additionally use intermediate features for supervision, and the following works~\cite{zagoruyko2016paying,tian2019contrastive,song2021tree} extract the deeper level of information in different aspects.
More recently, mutual learning~\cite{zhang2018-mutual}, a variant of KD, has attracted many interests as all the models~(students) are learning from each other simultaneously.
This practical learning strategy has been applied to person re-identification~\cite{fan2021-person, wang2020-person2}, object detection~\cite{wu2019-MLM}, and face recognition~\cite{deng2019-mutualface, banerjee2018-mutualface2}.
Apart from the applications of mutual learning, some researchers focus on improving mutual learning by introducing more supervision like the intermediate features~\cite{yao2020-mutualf1} or feature fusion~\cite{kim2021-mutualf2}.
Inspired by this, we propose utilizing an agent CNN optimized jointly with the ViT. The hard-coded inductive biases are transferred to the ViT under the mutual learning framework with adaptive intermediate feature supervision.

%% file: main/method.tex
\input{floats/figure_method_large.tex}

In this section, we first introduce the preliminaries of CNNs and ViTs. Then, based on the relationship of convolution layers and MHSA layers, agent CNNs are proposed to help train ViTs. Finally, we delineate the bootstrapping optimization algorithm where the agent and ViT are jointly optimized without pre-training. The workflow of our method is demonstrated in~\cref{fig:method_large}.

\subsection{Preliminaries}
\subsubsection{Convolution}
Convolution is the central part of CNNs, which accepts a two-dimensional feature map. For the sake of future discussion, we formularize the forward propagation with a sequence of visual tokens $X=(x_1, \ldots, x_n) \in \mathbb{R}^{n \times d_\text{in}}$ as its input, each of which is a $d_\text{in}$-dimensional embedded vector. Thus, the output sequence of convolution with kernel size $(k_h, k_w)$ is the sum of linear projection of $X$:
\begin{equation}
    Y_C = \sum_{i=1}^{N} \Phi_i X W_i \text{,}
\label{eq:cnn}
\end{equation}
where $\Phi_i$ is a constant sparse matrix representing the hard-coded inductive biases of localized dependencies, the size of receptive field $N=k_h \times k_w$, and the projection matrix $W_i \in \mathbb{R}^{d_\text{in} \times d_\text{out}}$ is trainable\footnotemark.

It is worth noting that the $1\times 1$ convolution has the form of $Y = XW$, which is equivalent to a fully connected~(FC) layer with the same projection matrix $W$.

\footnotetext{For simplicity, the bias terms are omitted. The detailed derivations are presented in~\cref{appendix:conv}.}

\subsubsection{MHSA}
The multi-head self-attention mechanism~(MHSA) in ViTs takes a sequence of visual tokens as its input and can also be formularized similar to~\cref{eq:cnn}:
\begin{equation}
    Y_M = \sum_{h=1}^H \Psi_h X W_h^{VO} \text{,}
\label{eq:mhsa}
\end{equation}
where $H$ is the number of heads, $W_h^{VO}=W_h^V W_h^O$ is the combination of two projection matrices~($W_h^V\in\mathbb{R}^{d \times d_k}$, $W_h^O\in\mathbb{R}^{d_k \times d}$, $d=H d_k$), and $\Psi_h\in\mathbb{R}^{n\times n}$ is the dense attention matrix computed based on the pair-wise similarity of linearly projected tokens.

\subsection{Agent CNN}
\input{floats/figure_method_cnn.tex}

Inspired by the similarity of~\cref{eq:cnn} and~\cref{eq:mhsa} that the convolution layer can be treated as a special case of MHSA layer with sparse relationship matrices $\Psi$, we propose constructing an \textit{agent} CNN based on a given ViT that will converge faster when trained from scratch.

\subsubsection{Generalized Convolution}
To begin with, we propose a generalized convolution layer in which the size of its receptive field $N$ equals to the head number $H$ of a MHSA layer, named as \textit{CONV}, with hard-coded inductive biases $\{\tilde{\Phi}_h\}_{h=1}^H$:
\begin{equation}
    Y_\text{CONV} = \sum_{h=1}^{H} \tilde{\Phi}_h X W_h \text{,}
\label{eq:g_conv}
\end{equation}
where $\tilde{\Phi}_1, \ldots, \tilde{\Phi}_H \in \mathbb{R}^{n\times n}$ are extracted from the set of hard-coded inductive biases $\{\Phi_1, \ldots, \Phi_{N'}\}$ of the $\lceil\sqrt{H}\rceil \times \lceil\sqrt{H}\rceil$ convolution~($N' = \lceil\sqrt{H}\rceil^2$) defined in~\cref{eq:cnn}.

\subsubsection{Constructing Agent CNN}
\label{sec:agent}
We start with a standard ViT model~(in~\cref{fig:method_cnn_a}) with $m$ encoder layers and finally build an agent CNN for introducing the inductive biases of CNNs.
\cref{fig:method_cnn_b} illustrates a base agent CNN by simply replacing the MHSA layers of ViT with CONV layers, which has introduced sparsity and localized biases.
Besides, the MLP in the agent are composed of two $1\times 1$ convolution layers which are equivalent to the fully connected layers in the FFNs of Transformers.

Moreover, as many preferred CNNs share the feature pyramid architecture~\cite{simonyan2014-vgg,he2016-deep,zagoruyko2016-wide,lin2017-feature} that the spacial size of feature maps shrunk as going deeper, we construct the final res-like agent CNN~(in~\cref{fig:method_cnn_c}) by:
(1) introducing a ResNet-style input projection block which contains two convolution layers and one max-pooling layer,
(2) adopting a configurable down-sample after each encoder layer.

With the hard-coded inductive biases, the agent can converge faster and with higher performance than training the corresponding ViT from scratch as shown in~\cref{fig:curv-c}.

\paragraph{Weight Sharing.}
Utilizing the homologous network architecture, our proposed agent accepts shared weights from the ViT model to reduce the training burden.
Due to the equivalence of $1\times 1$ convolution and FC layers, the FFNs in each encoder block of ViT can be directly shared by the agent.
Furthermore, when shared with the output projection $W_h^{VO}$ of MHSA in~\cref{eq:mhsa}, the CONV has the form of
\begin{equation}
    Y_\text{CONV} = \sum_{h=1}^H \tilde{\Phi}_h X W_h^{VO} \text{.}
\label{eq:shared_mhc}
\end{equation}

Let $y_c$ and $\tilde{y}_c$ be the $c$-th token of the output of MHSA and shared CONV accordingly. With the assumption that the input sequences are the same, denoted by $X$, the difference $y_\text{err}=y_c - \tilde{y}_c$ can be written as
\begin{equation}
    y_\text{err} = \sum_{h=1}^H (\tilde{\phi}_h - \psi_h) X W_h^{VO} \text{,}
\label{eq:y_err}
\end{equation}
where $\tilde{\phi}_h$ and $\psi_h$ are the $c$-th row of matrix $\tilde{\Phi}_h$ and $\Psi_h$ respectively. As there are no more than one non-zero element in $\tilde{\phi}_h$~(proved in~\cref{appendix:conv}), we can minimize the magnitude of $y_\text{err}$ by learning sparse and localized dependencies of matrix $\Psi_h$.

\subsection{Bootstrapping Optimization}
In this section, we describe how to jointly optimize the ViT and the agent by introducing the optimization objective and proposed training strategy.

\subsubsection{Adaptive Intermediate Supervision}
\label{sec:adaptive_supervision}
To inject the inductive biases of the agent into ViT without modifying ViT's architecture, we propose the adaptive intermediate supervision where the adapted feature maps of the agent supervise the corresponding visual sequences of the ViT. Let $F_A^{(\ell)}$ and $F_V^{(\ell)}$ denote the flattened feature maps and visual sequence of the $\ell$-th encoder layer of the agent and ViT in respective.
The adaptive intermediate loss of $\ell$-th layer of the ViT and agent is defined as
\begin{equation}
    \mathcal{L}_\text{feat}^{(\ell)} = \left\| \frac{\tilde{F}_A^{(\ell)}} {\| \tilde{F}_A^{(\ell)} \|_2} - \frac{F_V^{(\ell)}} {\| F_V^{(\ell)} \|_2} \right\|_2^2 \text{,}
\label{eq:feat_supervise}
\end{equation}
where $\tilde{F} := \mathrm{Adapt}(F)$ is the adapted feature, obtained from sequence interpolation or 2-dimensional average pooling. We have compared the different adaptive approaches in~\cref{exp:ablation}.

Finally, the adaptive intermediate supervision is the sum of all assigned layers $\varLambda$:
\begin{equation}
    \mathcal{L}_\text{feat} = \sum_{\ell\in\varLambda} \mathcal{L}_\text{feat}^{(\ell)} \text{.}
\label{eq:all_feat}
\end{equation}

\subsubsection{Optimization Objective}
\label{sec:optimization_obj}
Except for intermediate supervision, we introduce the mutual learning framework~\cite{zhang2018-mutual} that the predicted probabilities by the ViT~(denoted as $p_V$) and the agent~(denoted as $p_A$) learn from each other as
\begin{equation}
    \mathcal{L}_\text{mutual} = \mathcal{L}_\text{KD}(p_V, \bar{p}_A, y; T) + \mathcal{L}_\text{KD}(p_A, \bar{p}_V, y; T) \text{,}
\label{eq:mutual}
\end{equation}
where $\bar{p}$ represents that the variable $p$ is treated as a constant vector, \ie, no gradient is computed with regard to the variables in the forward propagation paths, $\mathcal{L}_\text{KD}$ is the knowledge distillation loss defined in~\cite{hinton2015-distilling} with temperature $T$, and $y$ denotes the ground truth label of the input image.

Above all, the optimization objective is summarized as
\begin{equation}
    \mathcal{L} = \alpha \mathcal{L}_\text{feat} + \beta \mathcal{L}_\text{mutual} \text{,}
\label{eq:final}
\end{equation}
and $\alpha$ and $\beta$ are the weighting hyperparameters for balancing the two terms.

\subsubsection{Bootstrapping Training Algorithm}
The bootstrapping training algorithm is given in~\cref{alg:1} where gradients computed from each network are aligned and jointly update the shared weights. The gradient alignment function $\mathrm{Align}(\nabla^A_S|\nabla^V_S)$ modifies the negative gradient direction from the agent as~\cite{yu2020-gradient}.

\input{floats/figure_multitask.tex}

\paragraph{Relationship with multi-task learning.} It is worth pointing out that bootstrapping learning is different from multi-task learning.
As shown in~\cref{fig:method_multitask}, the multi-task model $E$ only accepts one input $X$, while for bootstrapping learning, inputs to $E$ and $E'$ are different.
Moreover, the layers $E$ and $E'$ in~\cref{fig:method_multitask_a} shares the same weights $\Theta$.
In our case, as we constrain the difference between inputs of each encoder layer by~\cref{eq:all_feat} and when $\|y_\text{err}\|$ is small enough, the bootstrapping learning will degenerate to multi-task learning.

\input{floats/algorithm.tex}

%% file: floats/figure_method_large.tex
\begin{figure*}[!t]
\centering%
    \includegraphics[width=\linewidth]{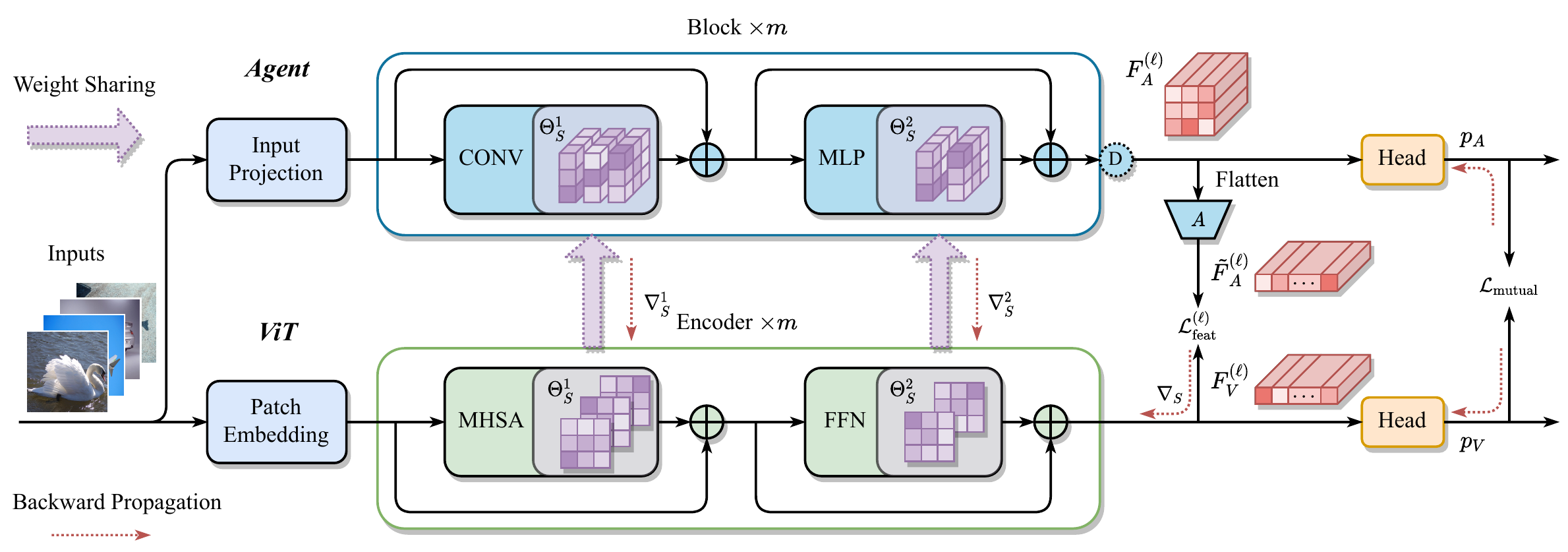}%
\caption{%
    Illustration of our proposed method for optimizing a vision Transformer along with an agent CNN from scratch.
    The agent CNN is constructed according to the ViT structure with inductive biases through generalized convolution~(CONV) and configurable down-sampling.
    The ViT learns agent's inductive biases from adaptive intermediate supervision $\mathcal{L}_\text{feat}^{(\ell)}$ and soft labels $\mathcal{L}_\text{mutual}$.
    Further, the weights of MHSA and FFN are shared to the agent CNN and trained by our proposed bootstrapping learning algorithm.
}%
\label{fig:method_large}%
\end{figure*}

%% file: floats/figure_method_cnn.tex
\begin{figure}[!t]
\centering%
\subfloat[Network architecture of ViT]{%
\includegraphics[width=\linewidth]{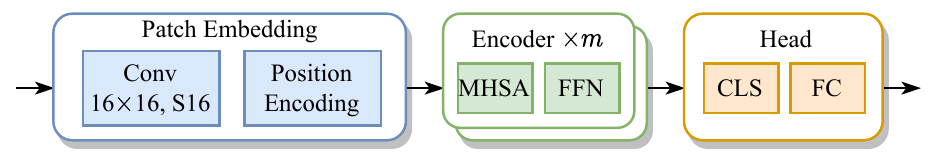}%
\label{fig:method_cnn_a}%
}\\
\subfloat[Network architecture of base agent CNN]{%
\includegraphics[width=\linewidth]{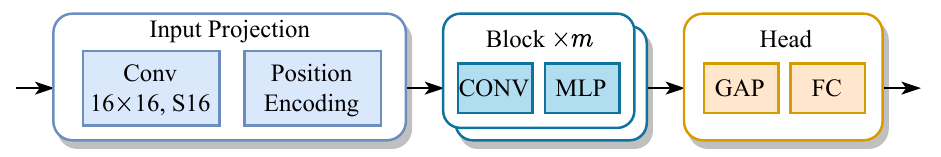}%
\label{fig:method_cnn_b}%
}\\
\subfloat[Network architecture of res-like agent CNN]{%
\includegraphics[width=\linewidth]{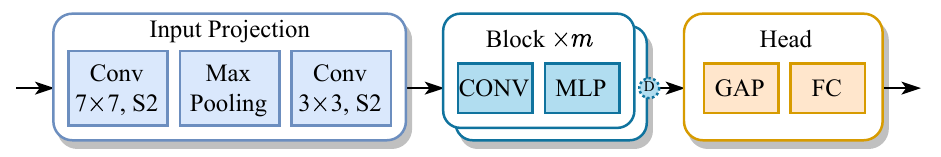}%
\label{fig:method_cnn_c}%
}
\caption{
    Illustration of the architecture of the ViT and our proposed agent CNN with base and res-like configurations. The generalized convolution~(CONV) substitutes for the MHSA, and the global average pooled feature replaces the CLS token in the ViT. Moreover, in the res-like agent, the feature pyramid is achieved by configurable down-sampling layers behind each block.
}%
\label{fig:method_cnn}%
\end{figure}

%% file: floats/figure_multitask.tex
\begin{figure}[!t]
\centering%
\subfloat[Bootstrapping Learning]{%
    \includegraphics[height=10em]{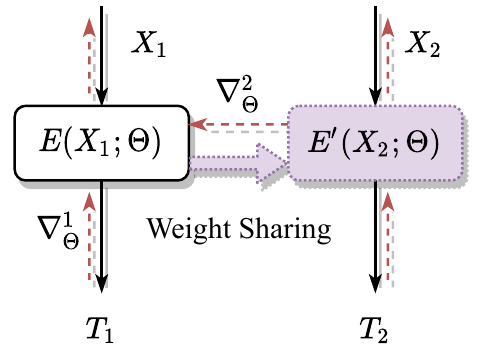}%
    \label{fig:method_multitask_a}%
}%
\subfloat[Multi-task Learning]{%
    \includegraphics[height=10em]{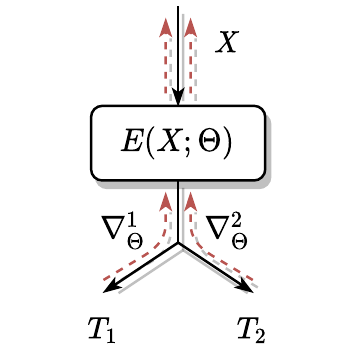}%
    \label{fig:method_multitask_b}%
}%
\caption{The relationship of our proposed bootstrapping learning and multi-task learning. Here, $T_1$ and $T_2$ are two different tasks~(optimization objective).}%
\label{fig:method_multitask}%
\end{figure}

%% file: floats/algorithm.tex
\begin{algorithm}[!t]
\caption{Bootstrapping optimizer for training the shared weights in FFN and MHSA layers in the ViT.}
\label{alg:1}
\begin{algorithmic}[1]
    \Require $\Theta_S$: the set of shared weights;
    $\Theta_V$, $\Theta_A$: the set of private weights in the ViT and agent respectively;
    $E_V^{(\ell)}(\cdot; \Theta_S, \Theta_V)$, $E_A^{(\ell)}(\cdot; \Theta_S, \Theta_A)$: the $\ell$-th encoder layer of the ViT and agent respectively;
    $\lambda$: learning rate.
    \While {\textbf{not} \textit{converge}}
        \State Compute the input feature map~(sequence) $X_V$, $X_A$ to the $\ell$-th encoder layer from the input image
        \State $Y_V \gets E_V^{(\ell)}(X_V; \Theta_S, \Theta_V)$
        \State $Y_A \gets E_A^{(\ell)}(X_A; \Theta_S, \Theta_A)$
        \State Compute the gradients $\nabla^V_S$, $\nabla^V_P$ \wrt $\Theta_S$ and $\Theta_V$ respectively ($\Theta_A$ keep as constant variables).
        \State Compute the gradients $\nabla^A_S$, $\nabla^A_P$ \wrt $\Theta_S$ and $\Theta_A$ respectively ($\Theta_V$ keep as constant variables).
        \State $\Theta_V \gets \Theta_V - \lambda \nabla^V_P$, $\Theta_A \gets \Theta_A - \lambda \nabla^A_P$
        \Comment update private weights
        \State $\Theta_S \gets \Theta_S - \nicefrac{\lambda}{2} (\nabla^V_S + \mathrm{Align}(\nabla^A_S|\nabla^V_S))$
        \Comment update shared weights in bootstrapping way
    \EndWhile
\end{algorithmic}
\end{algorithm}

%% file: main/experiments.tex
\subsection{Implementation}
\label{sec:implementation}
\input{floats/table-cifar.tex}
\paragraph{Datasets.}
Three widely-used image classification datasets are adopted to evaluate our proposed method as benchmarks, including CIFAR-10~\cite{krizhevsky2009-cifar}, CIFAR-100~\cite{krizhevsky2009-cifar}, and ImageNet-1k~\cite{krizhevsky2012-imagenet}. In particular, to simulate lack-of-data circumstances, 1\%, 5\%, and 10\% labeled samples are randomly extracted from the training partition of the ImageNet dataset.
Even though ViTs require strong data augmentations in previous approaches~\cite{dosovitskiy2021-vit,touvron2020-deit,yuan2021-tokenvit,wu2021-cvt}, in our implementation, both CNNs and ViTs are only optimized with several simple augmentation methods, including random resized cropping and random horizontal flipping.

\paragraph{Vision Transformers.}
We follow the network architectures introduced in~\cite{dosovitskiy2021-vit} and~\cite{touvron2020-deit} with slight modifications to the head number and embedding dimensions.
The detailed settings of ViTs are presented in~\cref{table:architecture}, in which ViT-S is a relatively small model with 6 layers, and the ViT-B is the same as DeiT~\cite{touvron2020-deit} with 12 layers.

\paragraph{Agent CNNs.}
The agents are constructed according to the given ViTs and thus share the same network settings as ViTs. Besides, detailed configurations of the base and res-like agents are described in~\cref{appendix:arch}.

\input{floats/table-architecture.tex}

\paragraph{Training Details and Selection of Hyperparameters.}
We implement our method with Pytorch~\cite{paszke2019-pytorch} framework.
AdamW~\cite{loshchilov2017-adamw} is used to optimize ViTs and agent CNNs in both standalone and joint training schemes, with the learning rate of $10^{-3}$ and the weight decay of $5\times 10^{-2}$.
However, conventional CNNs, such as ResNet~\cite{he2016-deep} and EfficientNet~\cite{tan2019-efficientnet}, are optimized by SGD~\cite{sutskever2013-importance} with the learning rate of $5\times 10^{-2}$ and the weight decay of $5\times 10^{-4}$.
We train all the settings for 240 epochs with the batch size of 32 on two Nvidia Tesla A100 GPUs. The cosine annealing is adopted as the learning rate decay schedule.

There are several hyperparameters involved in our method, including $\alpha$ and $\beta$ in~\cref{eq:final} and the temperature $T$ for knowledge distillation loss in~\cref{eq:mutual}. We set $\alpha=1$, $\beta=10$ and $T=4$ as default values and the sensitivity analysis is described in~\cref{exp:sensity}.

Finally, we set the intermediate feature supervisions decay linearly for preserving the capacity of the ViTs, and different decay strategies are compared in~\cref{exp:ablation}.

\subsection{Experimental Results}
\label{exp:results}

We evaluate our proposed method with the following comparison settings:
\begin{itemize}
    \item \textit{CNNs:} the standalone agents and traditional CNNs, \eg, EfficientNet-B2 and ResNet50.
    \item \textit{ViTs}: the original ViTs and variants for training with high efficiency, \eg, ViT-Sparse~\cite{correia2019-adaptively} and ViT-SAM~\cite{chen2021-vit-outperform}. All of them are trained from scratch.
    \item \textit{Pre-trained ViTs}: vision Transformers which have been pre-trained on ImageNet-1k and then fine-tuned to evaluation datasets.
    \item \textit{Ours Joint:} the vision Transformers and the agent are jointly optimized without weight sharing.
    \item \textit{Ours Shared:} the vision Transformers and the agent are jointly optimized with weight sharing.
\end{itemize}

\paragraph{Performance on CIFAR Datasets.}
\input{floats/figure_acc_curv.tex}

The comparison results on CIFAR-10/100 are shown in~\cref{table:result-cifar}, where the top-1 accuracy, number of parameters, and FLOPs are presented for each setting.
The discoveries are itemized as follows.
(1) When the agent CNNs are trained alone, the performance can hardly surpass the traditional CNNs, even if most of the inductive biases are hard-coded into the agents.
In~\cref{sec:agent-choice}, we discuss more about the choice of agents.
(2) Without pre-training or strong data augmentations, the ViTs perform terribly due to the dense connection of MHSA layers. Although Chen~\etal~\cite{chen2021-vit-outperform} have shown that ViTs can be optimized with SAM optimizer,  librated from the pre-training on mid-scale datasets, such as ImageNet-1k, it may not be the best choice for small datasets like CIFAR.
(3) Our proposed method has significantly outperformed the baseline settings, including the original ViTs and the variations. Particularly, ViT-S surpasses the original baseline by 7.82\% on CIFAR-10 and 14.49\% on CIFAR-100, which outperforms the agent and EfficientNet-B2 with fewer parameters.
Besides, both the ViTs and agents have benefited from our proposed method; however, with the help of the global receptive field, ViTs perform better.
(4) In the `Shared' settings, the bootstrapping learning strategy has shown that the shared weights can be optimized robustly with a limited deterioration of accuracy.
Such results are encouraging that the weights of ViTs can be directly transferred to a framework of hard-coded inductive biases so that ViTs can utilize the inductive biases without an extra set of parameters or sophisticated modifications.

Moreover, we plot the accuracy learning curves of our method along with the baseline settings in~\cref{fig:acc-curv}. It clearly illustrates that ViTs can converge as fast as CNNs and finally achieve the higher upper bound than CNNs'.

\paragraph{Performance on ImageNet.}
\input{floats/table-imagenet.tex}
The comparison results on ImageNet-1k with different amounts of labeled images are shown in~\cref{table:result-imagenet}, where 5\%, 10\%, and 50\% of the training images are randomly selected.
The conclusions of the CIFAR-10/100 dataset still hold in the ImageNet.
Particularly, the improvement of our method is prominent when data is extremely scarce, while others have shown inconspicuous amelioration or even impairment.

\subsection{Ablation Study}
\label{exp:ablation}
\input{floats/table-ablation-term.tex}

In this section, we go deep into our proposed method to figure out the function of loss terms~(introduced in~\cref{sec:optimization_obj}), the effects of different decay strategies~(in~\cref{sec:implementation}), and adaptive functions~(in~\cref{sec:adaptive_supervision}). The results of ViT-S are reported on CIFAR-100 dataset.

\subsubsection{Ablation of Loss Terms}
As delineated in~\cref{sec:optimization_obj}, the ultimate optimization objective has two terms: the adaptive intermediate supervision $\mathcal{L}_\text{feat}$ and the mutual learning term $\mathcal{L}_\text{mutual}$.
We evaluate the joint learning settings when supervised with the two terms separately in~\cref{table:ablation_term}, showing that both loss terms have contributed to the final result.
Particularly, $\mathcal{L}_\text{feat}$ boosts the accuracy by 13.69 percentage points.
Additionally, learning curves are plotted in~\cref{fig:curv-c} for better illustration. We can observe that the `Feat' term produces a more competitive result as it converges significantly faster than using the `Mutual' term.
Therefore, the supervision through the intermediate features has successfully injected the inductive biases into the ViT.

\subsubsection{Ablation of Decay Strategy}
The influence of the feature supervision decay strategy is shown in~\cref{table:ablation_term}. Without the decay strategy, the performance has declined by 1.35\%.
It can be explained that the constant supervision with inductive biases has constrained the ViTs from learning the long-range dependencies, and consequently, impaired the upper-bound of ViTs.

\subsubsection{Ablation of Adaptive Functions}
We evaluate our method with two intermediate feature adaptive functions: 1D sequence interpolation~(by default) and 2D average pooling. The comparison results are shown in~\cref{table:ablation_term}, in which the sequence interpolation is superior to the average pooling.

\subsection{The Choice of Agent CNNs}
\label{sec:agent-choice}
\input{floats/table-cnn_arch.tex}
In~\cref{sec:agent}, we have introduced agent CNNs with two different network architectures~(base and res-like). \cref{table:cnn_arch} shows that the performances with the res-like configuration are uniformly better than the base agent.

\subsection{Sensitivity Analysis of Hyperparameters}
\label{exp:sensity}
\input{floats/figure_sensity.tex}

The sensitivity analysis of the hyperparameters in our methods is illustrated in~\cref{fig:sensity}, including $\alpha$, $\beta$ in~\cref{eq:final}, and $T$ in~\cref{eq:mutual}.
It shows that our method is robust to the variation of $\alpha$ and $T$. However, $\beta$ has a more significant influence, as the ViTs perform better when more inductive biases are used to supervise the ViTs.

%% file: floats/table-cifar.tex
\begin{table*}[!t]
\centering
\setlength{\tabcolsep}{8pt}
\caption{Comparison results on CIFAR-10 and CIFAR-100. The top-1 accuracy, number of parameters, and FLOPs are reported separately. `$\dagger$' indicates that the initial weights of pre-trained ViT-B are acquired from the official repository of DeiT. The comparison settings are classified in the `Model' column. The values in blue color indicates top-1 accuracy improvements compared with the corresponding ViT trained from scratch.}
\begin{tabular}{ll *{3}{r} *{3}{r}}
    \toprule
    \multicolumn{1}{l}{\multirow{2}{*}{\textbf{Model}}}
    & \multicolumn{1}{l}{\multirow{2}{*}{\textbf{Method}}}
    & \multicolumn{3}{c}{\textbf{CIFAR-10}}
    & \multicolumn{3}{c}{\textbf{CIFAR-100}} \\
    \cmidrule(lr){3-5} \cmidrule(lr){6-8}
    & & \multicolumn{1}{c}{\textbf{Acc}}
    & \multicolumn{1}{c}{\textbf{\#param.}}
    & \multicolumn{1}{c}{\textbf{FLOPs}}
    & \multicolumn{1}{c}{\textbf{Acc}}
    & \multicolumn{1}{c}{\textbf{\#param.}}
    & \multicolumn{1}{c}{\textbf{FLOPs}} \\
    \midrule

    \multirow{4}{*}{CNNs}
    & EfficientNet-B2 & 94.14 & 7.71M & 0.70G & 75.55 & 7.84M & 0.70G \\
    & ResNet50 & 94.92 & 23.53M & 4.14G & 77.57 & 23.71M & 4.14G \\
    & Agent-S & 94.18 & 8.66M & 3.37G & 74.62 & 8.73M & 3.37G \\
    & Agent-B & 94.83 & 25.05M & 9.46G & 74.78 & 25.91M & 9.46G \\
    \midrule

    \multirow{6}{*}{ViTs}
    & ViT-S & 87.32 & 6.28M & 1.37G & 61.25 & 6.30M & 1.37G \\
    & ViT-S-SAM & 87.77 & 6.28M & 1.37G & 62.60 & 6.30M & 1.37G \\
    & ViT-S-Sparse & 87.43 & 6.28M & 1.37G & 62.29 & 6.30M & 1.37G \\
    & ViT-B & 79.24 & 21.67M & 4.62G & 53.07 & 21.70M & 4.62G \\
    & ViT-B-SAM & 86.57 & 21.67M & 4.62G & 58.18 & 21.70M & 4.62G \\
    & ViT-B-Sparse & 83.87 & 21.67M & 4.62G & 57.22 & 21.70M & 4.62G \\
    \midrule

    \multirow{2}{1.8cm}{Pre-trained ViTs}
    & ViT-S & 95.70 & 6.28M & 1.37G & 80.91 & 6.30M & 1.37G \\
    & ViT-B$^\dagger$ & 97.17 & 21.67M & 4.62G & 84.95 & 21.70M & 4.62G \\
    \midrule

    \multirow{4}{1.8cm}{Ours Joint}
    & Agent-S & 94.90 & 8.66M & 3.37G & 74.06 & 8.73M & 3.37G \\
    & ViT-S & 95.14~(\Blue{+7.82}) & 6.28M & 1.37G & 76.19~(\Blue{+14.94}) & 6.30M & 1.37G \\
    & Agent-B & 95.06 & 25.05M & 9.46G & 76.57 & 25.91M & 9.46G \\
    & ViT-B & 95.00~(\Blue{+15.76}) & 21.67M & 4.62G & 77.83~(\Blue{+24.76}) & 21.70M & 4.62G \\
    \midrule

    \multirow{4}{1.8cm}{Ours Shared}
    & Agent-S & 93.22 & \multicolumn{1}{c}{-} & \multicolumn{1}{c}{-}
    & 74.06 & \multicolumn{1}{c}{-} & \multicolumn{1}{c}{-} \\

    & ViT-S & 93.72~(\Blue{+6.40}) & 6.28M & 1.37G & 75.50~(\Blue{+14.25}) & 6.30M & 1.37G \\
    & Agent-B & 92.66 & \multicolumn{1}{c}{-} & \multicolumn{1}{c}{-}
    & 74.11 & \multicolumn{1}{c}{-} & \multicolumn{1}{c}{-} \\

    & ViT-B & 93.34~(\Blue{+14.10}) & 21.67M & 4.62G & 75.71~(\Blue{+22.64}) & 21.70M & 4.62G \\
    \bottomrule
\end{tabular}%
\label{table:result-cifar}
\end{table*}

%% file: floats/table-architecture.tex
\begin{table}[!t]
\centering
\setlength{\tabcolsep}{5pt}
\caption{Detailed configurations of ViTs and agent CNNs.}
\begin{tabular}{l cccc}
    \toprule
    & \multicolumn{1}{c}{\textbf{ViT-S}}
    & \multicolumn{1}{c}{\textbf{ViT-B}}
    & \multicolumn{1}{c}{\textbf{Agent-S}}
    & \multicolumn{1}{c}{\textbf{Agent-B}} \\
    \midrule
    Layers~($m$) & $6$ & $12$ & $6$ & $12$ \\
    Hidden Size~($d$) & $288$ & $384$ & $288$ & $384$ \\
    Heads~($H$) & $9$ & $6$ & $9$ & $6$ \\
    \bottomrule
\end{tabular}%
\label{table:architecture}
\end{table}

%% file: floats/figure_acc_curv.tex
\begin{figure*}[!t]%
\centering%
\subfloat[Accuracy learning curves on CIFAR-10.]{%
    \includegraphics[width=0.33\linewidth]{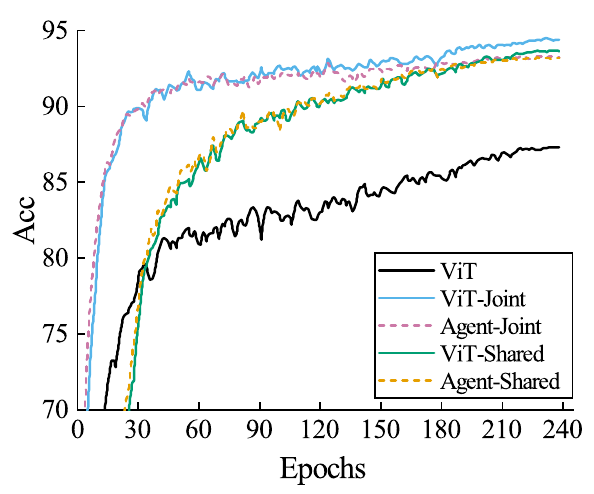}%
\label{fig:curv-a}%
}%
\subfloat[Accuracy learning curves on CIFAR-100.]{%
    \includegraphics[width=0.33\linewidth]{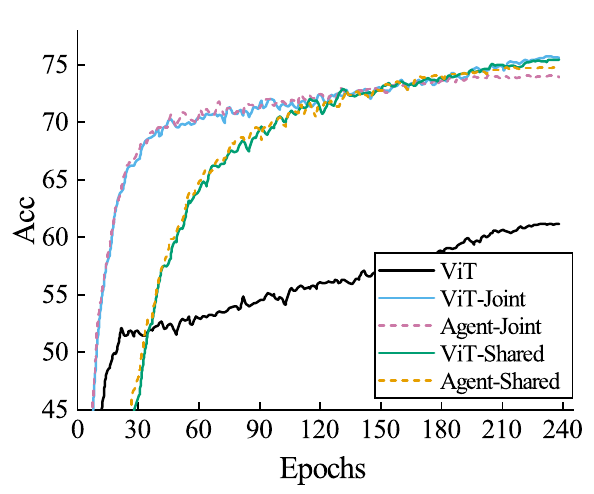}%
\label{fig:curv-b}%
}
\subfloat[Ablation study on CIFAR-100.]{%
    \includegraphics[width=0.33\linewidth]{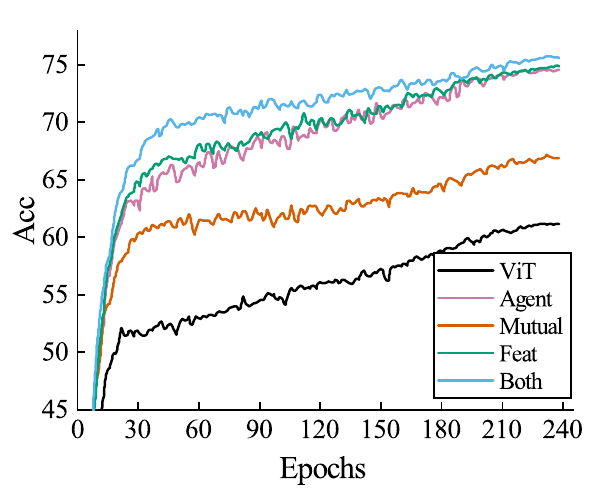}%
\label{fig:curv-c}%
}%
\caption{Accuracy learning curves of our proposed method and baseline settings on CIFAR-10 and CIFAR-100 datasets.
Specifically, We compare the accuracy of training ViT from scratch and training jointly with the agent CNN. Here, `Shared' and `Joint' represent jointly training both models with and without weight sharing, respectively.
Besides, we compare the results of the ablation study of loss terms in (c), where `Mutual' means training with mutual knowledge distillation term $\mathcal{L}_\text{mutual}$ only, and `Feat' denotes training with adaptive intermediate supervision $\mathcal{L}_\text{feat}$ only. In addition, the curve of training the agent model alone is plotted as `Agent'.
}%
\label{fig:acc-curv}%
\end{figure*}

%% file: floats/table-imagenet.tex
\begin{table}[!t]
\centering
\setlength{\tabcolsep}{4pt}
\caption{Comparison results on ImageNet-1k with 5\%, 10\%, and 50\% annotated samples.}
\resizebox{\linewidth}{!}{\begin{tabular}{l *{3}{r}}
    \toprule
    \multicolumn{1}{l}{\textbf{Mehod}}
    & \multicolumn{1}{c}{\textbf{5\% images}}
    & \multicolumn{1}{c}{\textbf{10\% images}}
    & \multicolumn{1}{c}{\textbf{50\% images}} \\
    \midrule
    ResNet50 & 35.43 & 50.86 & 70.05 \\
    Agent-B & 35.28 & 47.46 & 68.13 \\
    \midrule
    ViT-B & 16.60 & 28.11 & 63.40 \\
    ViT-B-SAM & 16.67 & 28.66 & 64.37 \\
    ViT-B-Sparse & 10.39 & 28.92 & 66.01 \\
    \midrule
    Ours-Joint & 36.01~(\Blue{+19.41}) & 49.73~(\Blue{+21.62}) & 71.36~(\Blue{+7.96}) \\
    Ours-Shared & 33.06~(\Blue{+16.46}) & 45.75~(\Blue{+17.64}) & 66.48~(\Blue{+3.08}) \\
    \bottomrule
\end{tabular}}%
\label{table:result-imagenet}
\end{table}

%% file: floats/table-ablation-term.tex
\begin{table}[!t]
\centering
\setlength{\tabcolsep}{4pt}
\caption{Ablation Study of the joint learning ViT-S method on CIFAR-100. In the column of `Feat', `\cmark' denotes using the default settings, `No Decay' represents the weight $\beta$ of the $\mathcal{L}_\text{feat}$ keep constant throughout the whole training process, and `AP-2D' means using 2D average pooling as adaptive function in~\cref{eq:feat_supervise}.}
\begin{tabular}{l rrr c}
    \toprule
    \multicolumn{1}{l}{\textbf{Settings}}
    & \multicolumn{1}{c}{\textbf{Mutual}}
    & \multicolumn{1}{c}{\textbf{Feat}}
    & \multicolumn{1}{c}{\textbf{Acc}}
    & \multicolumn{1}{c}{\textbf{Training Time}} \\
    \midrule
    \multirow{1}{*}{Baseline}
    & \xmark & \xmark & 61.25 & 3.3h\\
    \midrule
    \multirow{1}{*}{KD Only}
    & \cmark & \xmark & 67.16 & 3.6h \\
    \midrule
    \multirow{3}{*}{Feat Only}
    & \xmark & No Decay & 73.59 & 3.8h \\
    & \xmark & \cmark & 74.94 & 3.8h \\
    & \xmark & AP-2D & 71.06 & 6.6h \\
    \midrule
    \multirow{2}{*}{Both}
    & \cmark & No Decay & 75.15 & 3.8h \\
    & \cmark & \cmark & 76.19 & 3.8h \\
    \bottomrule
\end{tabular}
\label{table:ablation_term}
\end{table}

%% file: floats/table-cnn_arch.tex
\begin{table}[!t]
\centering
\caption{Joint training results on CIFAR-100 when using the agent with different network architectures.}
\begin{tabular}{lccc}
    \toprule
    \textbf{Acc}
    & \textbf{Agent-S} & \textbf{ViT-S-Joint} & \textbf{Agent-S-Joint} \\
    \midrule
    Base & 72.73 & 73.18 & 73.79 \\
    Res-like & 74.78 & 76.19 & 74.06 \\
    \bottomrule
\end{tabular}%
\label{table:cnn_arch}
\end{table}

%% file: floats/figure_sensity.tex
\begin{figure}[!t]
\centering%
\includegraphics[width=1.0\linewidth]{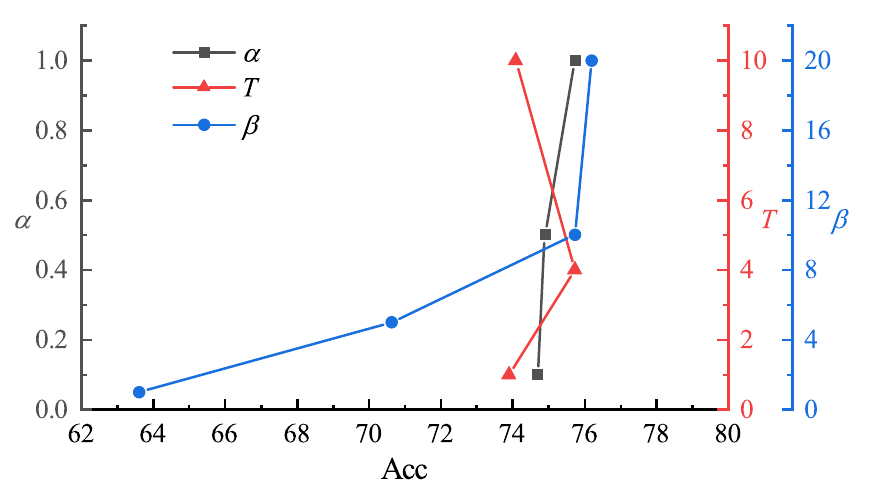}%
\caption{
    Sensitivity analysis of hyperparameters. The results of bootstrapping learning the ViT-S are reported on CIFAR-100.
}%
\label{fig:sensity}%
\end{figure}

%% file: main/conclusion.tex
In this paper, we propose to dissolve optimizing vision Transformers~(ViTs) efficiently without pre-training or strong data augmentations.
Our goal is to introduce inductive biases from convolutional neural networks~(CNNs) to ViTs while preserving the network architecture of ViTs for higher upper bound, and further setting up more suitable optimization objectives.
To this end, we propose to optimize the ViT jointly with an agent CNN constructed corresponding to the network architecture of the ViT.
The ViT learns inductive biases through adaptive intermediate supervision and predicted probabilities. Besides, a bootstrapping training algorithm is proposed to optimize both the ViT and agent with weight sharing.
Extensive experiments have shown encouraging results that the inductive biases help ViTs converge significantly faster and outperform conventional CNNs with fewer parameters.
In future work, we will extend our method beyond CNN-style inductive biases and introduce more interpretable features to ViTs.

%% file: appendices/conv.tex
In this section, we formularize convolution in the matrix form consistent with MHSA.
Let $\bm{F}^\text{in}\in\mathbb{R}^{H\times W\times d_\text{in}}$ denote a 2D input feature map to a $k_h \times k_w$ convolution layer. The receptive field $N$ of the convolution layer is defined as $N=k_h \times k_w$.
For the sake of simplicity, we assume that the input feature and the output feature share the same size, \ie, $\bm{F}^\text{out} \in\mathbb{R}^{H\times W\times d_\text{out}}$, and the padding value of the convolution is set to zero.
The output feature map at position $(h, w)$ only depends on the neighborhood $\Delta_{h,w} = \{(h_i, w_j)\}_{(i, j)\in [k_h]\times[k_w]}$ of the input feature map:
\begin{equation}
    \bm{F}_{h,w}^\text{out} = \sum_{(i,j)\in [k_h]\times[k_w]} \Delta_{h,w}[\bm{F}^\text{in}]_{i,j} \bm{W}_{i,j} \text{,}
    \label{eq:1}
\end{equation}
where operation $\Delta_{h,w}[\cdot]$ denotes the selection of neighborhood of input 2D feature by the neighborhood indices, $\bm{W}_{i,j}\in\mathbb{R}^{d_\text{in}\times d_\text{out}}$ denotes the linear projection matrix, and $\bm{W}\in\mathbb{R}^{k_h\times k_w\times d_\text{in}\times d_\text{out}}$ is the parameter tensor of the convolution layer $\mathrm{Conv}(\cdot; \bm{W})$.

Similarly, as we flatten the input feature map to 1D visual sequence $X=\mathrm{Flatten}(\bm{F}^\text{in}) \in \mathbb{R}^{n\times d_\text{in}}$, $n=HW$, the $q$-th output visual token $y_q\in\mathbb{R}^{d_\text{out}}$ can be calculated by
\begin{equation}
    y_q = \sum_{i=1}^{|\Delta_q|} \Delta_q[X]_i \bm{W}_i \text{,}
\end{equation}
where $\Delta_q = \{p_1, \ldots, p_N\}$ is the sequence of 1D coordinates according to flattened indices $\Delta_{h,w}$, $\Delta_q[X]$ is the extracted subsequence of $X$, and $\bm{W}_i$ is the linear projection matrix corresponding to $\bm{W}_{i,j}$ in~\cref{eq:1}.
Therefore, the output sequence $Y\in\mathbb{R}^{n\times d_\text{out}}$ is the stack of output tokens:
\begin{equation}
    Y = \begin{bmatrix}
        y_1 \\ \vdots \\ y_n
    \end{bmatrix}
     = \sum_{i=1}^{N}\begin{bmatrix}
        \Delta_1[X]_i \\
        \vdots \\
        \Delta_n[X]_i
     \end{bmatrix} \bm{W}_i \text{.}
\end{equation}

Here, we abuse some notations for further derivation by setting $\Delta_i[X] = \begin{bmatrix}\Delta_1[X]_i^\top & \cdots & \Delta_n[X]_i^\top \end{bmatrix}^\top$ as the shift of input sequence and $W_i := \bm{W}_i$.
It is worth noting that $\Delta_q[X]_i$ is the selection of one input visual token and can be treated as a linear transformation matrix $\phi_q^{(i)}\in\mathbb{R}^{1\times n}$.
Specifically,
\begin{equation}
    \phi_{q,p}^{(i)} = \begin{cases}
        1\text{,} & \mbox{if~} p=p_i=\Delta_q^{(i)} \text{,}\\
        0\text{,} & \mbox{otherwise.}
    \end{cases}
\end{equation}
Hence, $\phi_q^{(i)}$ has at most one non-zero element~($\phi_q^{(i)}=\bm{0}$ when $p_i$ is the zero padding index).

As such, the output sequence can be simplified as
\begin{equation}
    Y = \sum_{i=1}^N \Phi_i X W_i \text{,}
\end{equation}
where
\begin{equation}
    \Phi_i = \begin{bmatrix}
        \phi_1^{(i)} \\
        \vdots \\
        \phi_n^{(i)}
    \end{bmatrix}
\end{equation}
is the constant matrix with hard-coded inductive biases.

%% file: appendices/interdimate_sup.tex
\begin{figure}[!t]
\centering%
\includegraphics[width=1.0\linewidth]{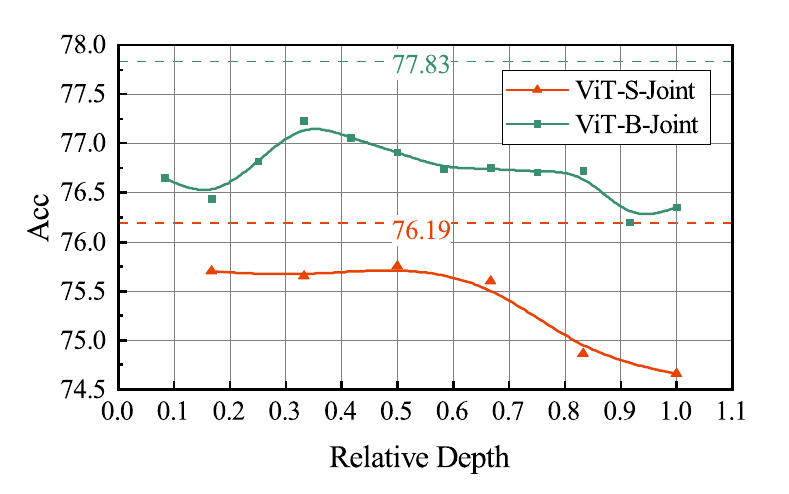}%
\caption{
    Accuracy variation when the intermediate supervision at different layers is removed.
}%
\label{fig:ablation-layers}%
\end{figure}

In this section, we analyze the effects when the intermediate supervisions act upon different layers.~\cref{fig:ablation-layers} shows the top-1 accuracy variation on CIFAR-100 dataset when the supervision at relative depth is removed from the final loss. The dash lines represent the performance without removing supervision for any layer.
For the small settings, the importance increases significantly as going deeper. However, the base model relies on both shallow and deep layers.

%% file: appendices/vis_mhsa.tex
\begin{wrapfigure}{r}{0.38\linewidth}%
\centering%
\subfloat[]{\includegraphics[width=0.48\linewidth]{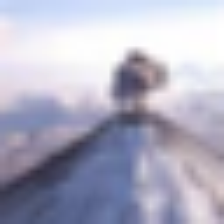}\label{fig:vis-img-1}}%
\hfill
\subfloat[]{\includegraphics[width=0.48\linewidth]{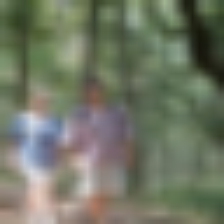}\label{fig:vis-img-2}}%
\caption{Selected images from CIFAR-100 for visualization.}%
\label{fig:vis-img}%
\end{wrapfigure}
The visualizations of learned self-attentions are shown in~\cref{fig:vis-mhsa-1} and~\cref{fig:vis-mhsa-2} for two randomly selected images from CIFAR-100.
The heat maps of layer 2, 4, and 6 reveal that the inductive biases, such as sparsity and localized relationship, have been injected into ViTs, especially for shallow layers.

%% file: appendices/vis_seq.tex
The visualizations of intermediate features for both agent CNN and the ViT are shown in~\cref{fig:vis-feat-1} and~\cref{fig:vis-feat-2} for two different input images~(\cref{fig:vis-img-1} and~\cref{fig:vis-img-2}).
As we can observe, (1) when trained independently, the CNNs tend to produce smooth features while ViTs tend to generate sharp features due to the global attention mechanism;
(2) when optimized jointly, as the inductive biases have been injected into the ViTs, they tend to pay more attention to the real object, which is similar to CNNs, and thus more robust to the disturbances.

%% file: appendices/arch.tex
\begin{table}[!t]
\centering
\setlength{\tabcolsep}{2pt}
\newcommand{\tabincell}[2]{\begin{tabular}{@{}#1@{}}#2\end{tabular}}
\caption{Network architectures of base agent CNNs.}
\resizebox{\linewidth}{!}{
\begin{tabular}{|c|c|c|}
    \hline
    \textbf{Layer} & \textbf{Agent-S} & \textbf{Agent-B} \\ \hline
    \makecell{Input \\ Projection}
    & 16$\times$16 Conv, 288, S16
    & 16$\times$16 Conv, 384, S16\\ \hline 

    \multirow{4}{*}{Blocks}
    & \multirow{4}{*}{$\begin{bmatrix}\text{H9, CONV, 288} \\ \text{1$\times$1 Conv, 1152} \\ \text{1$\times$1 Conv, 288}\\ \end{bmatrix}\times$6}
    & \multirow{4}{*}{$\begin{bmatrix}\text{H6, CONV, 384} \\ \text{1$\times$1 Conv, 1536} \\ \text{1$\times$1 Conv, 384}\\ \end{bmatrix}\times$12} \\
    & & \\
    & & \\
    & & \\ \hline
    \multirow{2}{*}{Head} & \multicolumn{2}{c|}{Global Average Pooling} \\
    & \multicolumn{2}{c|}{FC} \\ \hline
\end{tabular}%
}
\label{table:appendix_base_agent}
\end{table}

\begin{table}[!t]
\centering
\setlength{\tabcolsep}{2pt}
\newcommand{\tabincell}[2]{\begin{tabular}{@{}#1@{}}#2\end{tabular}}
\caption{Network architectures of res-like agent CNNs.}
\resizebox{\linewidth}{!}{
\begin{tabular}{|c|c|c|}
    \hline
    \textbf{Layer} & \textbf{Agent-S} & \textbf{Agent-B} \\ \hline
    \makecell{Input \\ Projection}
    & \makecell{7$\times$7 Conv, 64, S2 \\2$\times$2, Max Pooling, S2 \\ 3$\times$3 Conv, 288, S2}
    & \makecell{7$\times$7 Conv, 64, S2 \\2$\times$2, Max Pooling, S2 \\ 3$\times$3 Conv, 384, S2}\\ \hline 

    \multirow{14}{*}{Blocks}
    & \multirow{4}{*}{$\begin{bmatrix}\text{H9, CONV, 288} \\ \text{1$\times$1 Conv, 1152} \\ \text{1$\times$1 Conv, 288}\\ \end{bmatrix}\times$1}
    & \multirow{4}{*}{$\begin{bmatrix}\text{H6, CONV, 384} \\ \text{1$\times$1 Conv, 1536} \\ \text{1$\times$1 Conv, 384}\\ \end{bmatrix}\times$2} \\
    & & \\
    & & \\
    & & \\ \cline{2-3}
    & Down-sampling & Down-sampling \\ \cline{2-3}

    & \multirow{4}{*}{$\begin{bmatrix}\text{H9, CONV, 288} \\ \text{1$\times$1 Conv, 1152} \\ \text{1$\times$1 Conv, 288}\\ \end{bmatrix}\times$1}
    & \multirow{4}{*}{$\begin{bmatrix}\text{H6, CONV, 384} \\ \text{1$\times$1 Conv, 1536} \\ \text{1$\times$1 Conv, 384}\\ \end{bmatrix}\times$2} \\
    & & \\
    & & \\
    & & \\ \cline{2-3}
    & Down-sampling & Down-sampling \\ \cline{2-3}

    & \multirow{4}{*}{$\begin{bmatrix}\text{H9, CONV, 288} \\ \text{1$\times$1 Conv, 1152} \\ \text{1$\times$1 Conv, 288}\\ \end{bmatrix}\times$4}
    & \multirow{4}{*}{$\begin{bmatrix}\text{H6, CONV, 384} \\ \text{1$\times$1 Conv, 1536} \\ \text{1$\times$1 Conv, 384}\\ \end{bmatrix}\times$8} \\
    & & \\
    & & \\
    & & \\ \hline

    \multirow{2}{*}{Head} & \multicolumn{2}{c|}{Global Average Pooling} \\
    & \multicolumn{2}{c|}{FC} \\ \hline
\end{tabular}%
}
\label{table:appendix_res_agent}
\end{table}

The network architectures of agent CNNs are given in~\cref{table:appendix_base_agent} and~\cref{table:appendix_res_agent}, respectively for the base agent CNN and the res-like agent CNN. Here, `H9' and `H6' denotes generalized convolution with receptive field of 9 and 6 respectively. The input images are resized to $224\times 224$ pixels for both agent CNNs and ViTs. In each block, the CONV layer replaces the MHSA layer in ViTs and the following MLP layer is composed of two 1$\times$1 convolution layers.
Finally, features from the global average pooling layer are fed into a fully connected~(FC) layer for classification.

%% file: appendices/visualization.tex
\begin{table*}[!p]
\centering
\setlength{\tabcolsep}{1pt}%
\caption{Visualization of average attention for input~\cref{fig:vis-img-1}.}%
\resizebox{\linewidth}{!}{
\begin{tabular}{l| *{3}{c}}%
    \toprule
    \textbf{Layer} & 2 & 4 & 6 \\ \midrule
    \textbf{\makecell{ViT-S}}
    & \makecell{\includegraphics[width=0.4\linewidth]{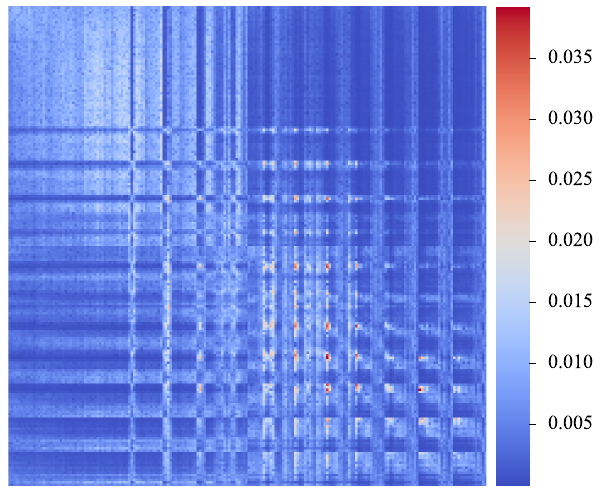}}%
    & \makecell{\includegraphics[width=0.4\linewidth]{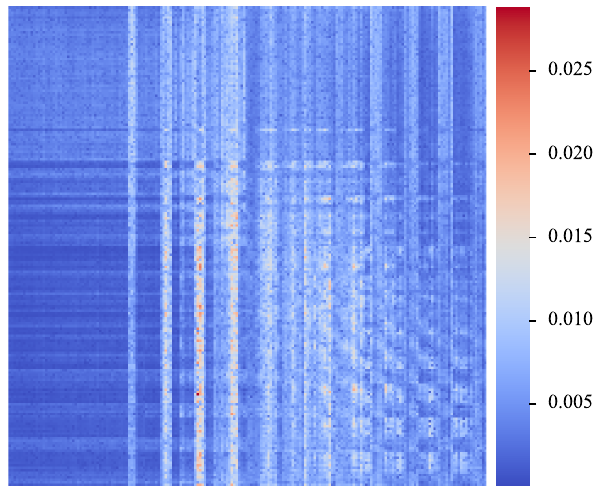}}%
    & \makecell{\includegraphics[width=0.4\linewidth]{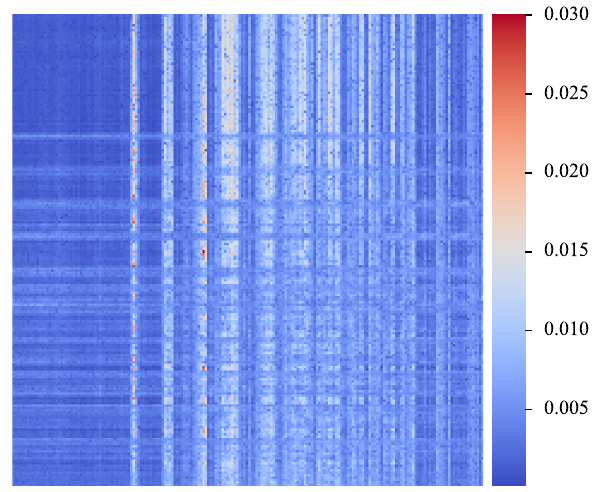}}%
    \\ \midrule

    \textbf{\makecell{Joint\\ViT-S}}
    & \makecell{\includegraphics[width=0.4\linewidth]{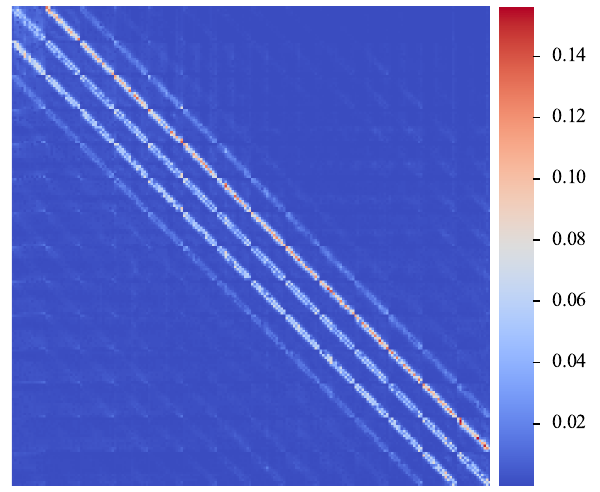}}%
    & \makecell{\includegraphics[width=0.4\linewidth]{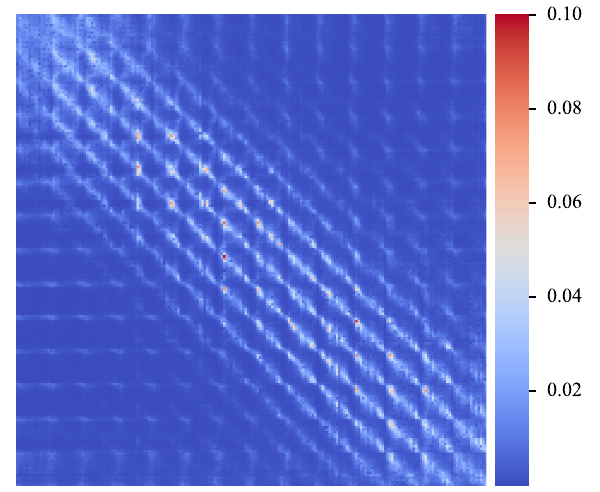}}%
    & \makecell{\includegraphics[width=0.4\linewidth]{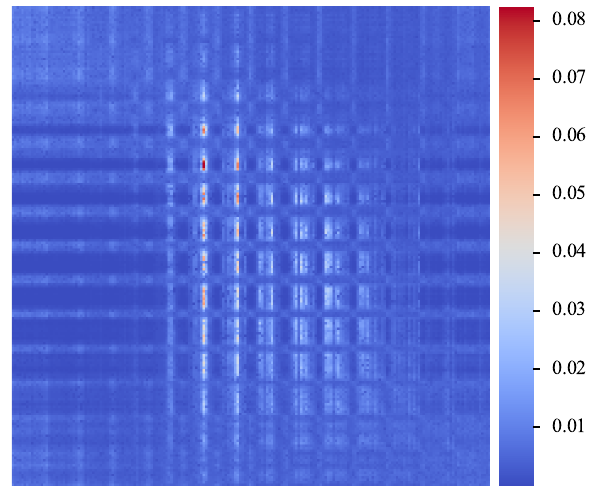}}%
    \\ \bottomrule
\end{tabular}%
}
\label{fig:vis-mhsa-1}%
\end{table*}
\begin{table*}[!tp]
\centering
\setlength{\tabcolsep}{1pt}%
\caption{Visualization of average attention for input~\cref{fig:vis-img-2}.}%
\resizebox{\linewidth}{!}{
\begin{tabular}{l| *{3}{c}}%
    \toprule
    \textbf{Layer} & 2 & 4 & 6 \\ \midrule
    \textbf{\makecell{ViT-S}}
    & \makecell{\includegraphics[width=0.4\linewidth]{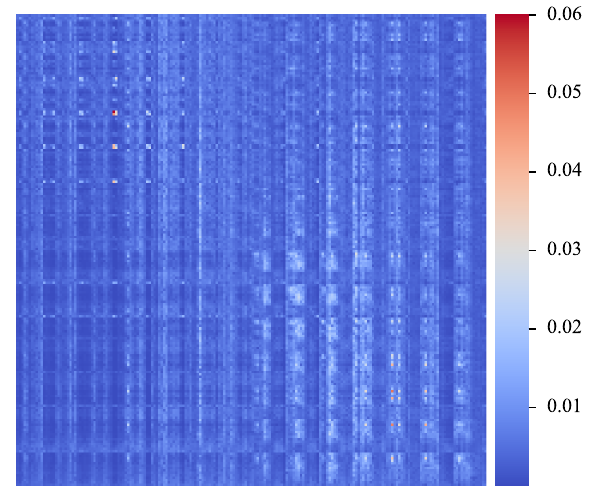}}%
    & \makecell{\includegraphics[width=0.4\linewidth]{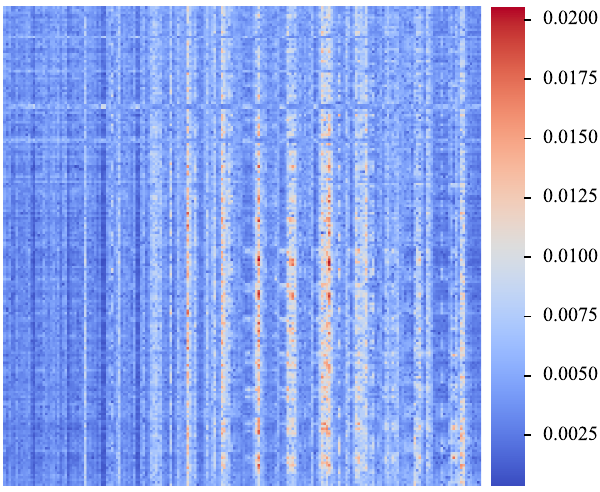}}%
    & \makecell{\includegraphics[width=0.4\linewidth]{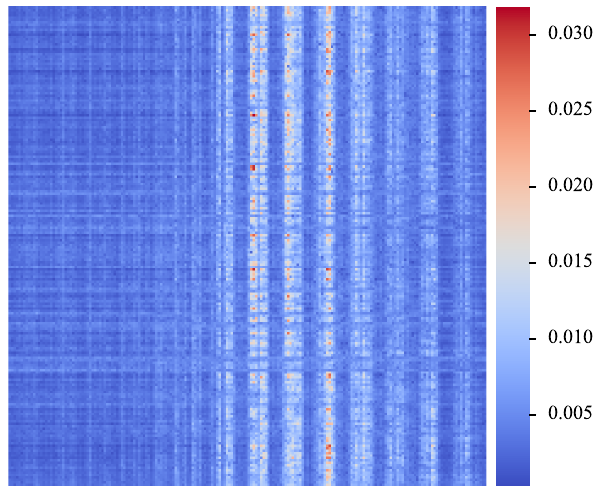}}%
    \\ \midrule

    \textbf{\makecell{Joint\\ViT-S}}
    & \makecell{\includegraphics[width=0.4\linewidth]{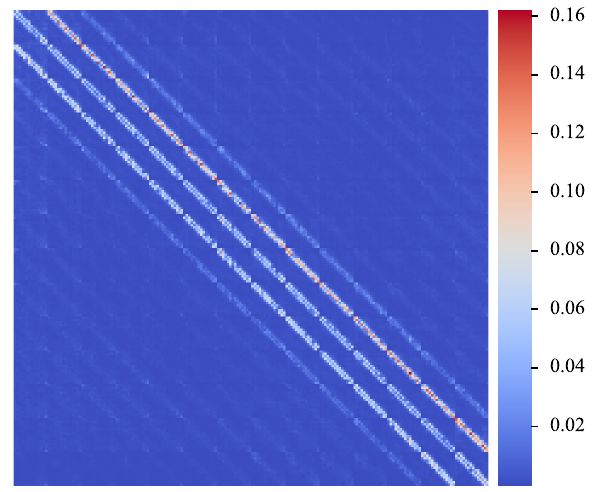}}%
    & \makecell{\includegraphics[width=0.4\linewidth]{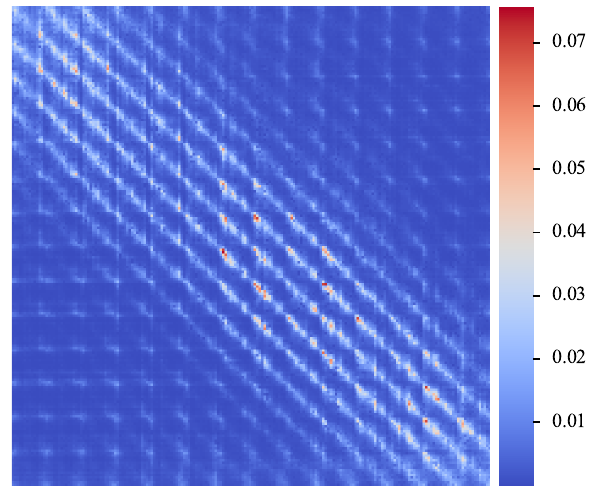}}%
    & \makecell{\includegraphics[width=0.4\linewidth]{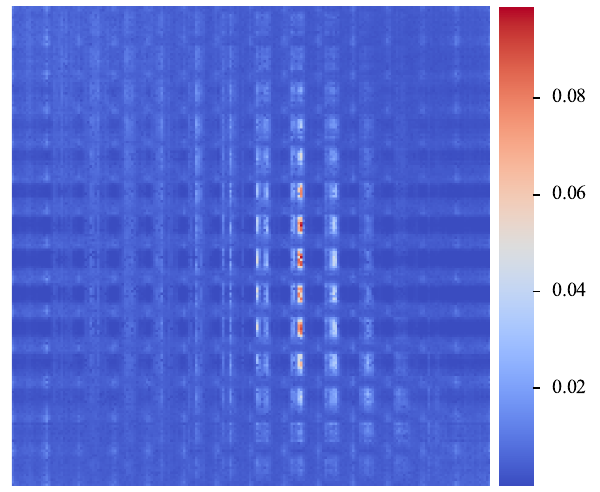}}%
    \\ \bottomrule
\end{tabular}%
}
\label{fig:vis-mhsa-2}%
\end{table*}

\clearpage
\begin{table*}[!p]
\centering
\setlength{\tabcolsep}{1pt}%
\caption{Visualization of intermediate features for input~\cref{fig:vis-img-1}. Please zoom for better view.}%
\resizebox{0.9\linewidth}{!}{
\begin{tabular}{l| cc|cc}%
    \toprule
    \textbf{Layer} & \textbf{Agent-S} & \textbf{ViT-S} & \textbf{Agent-S-Joint} & \textbf{ViT-S-Joint} \\ \midrule
    \makecell{2}
    & \makecell{\includegraphics[width=0.25\linewidth]{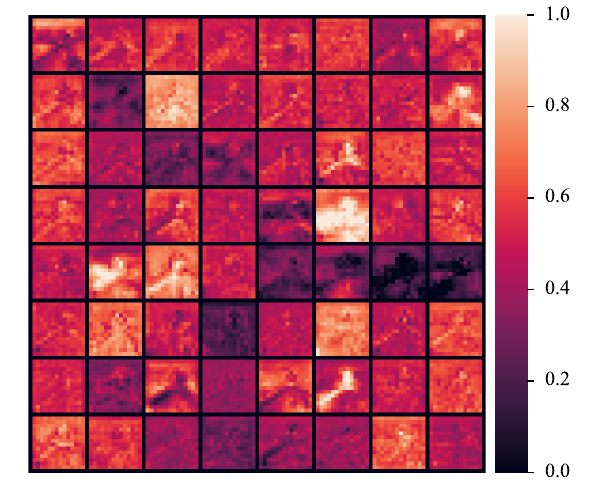}}%
    & \makecell{\includegraphics[width=0.25\linewidth]{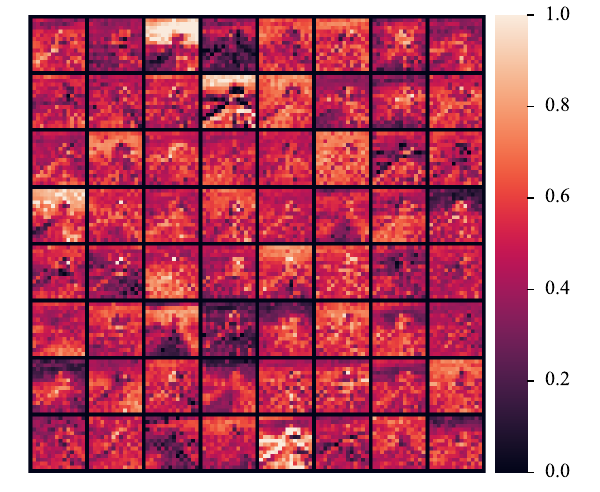}}%
    & \makecell{\includegraphics[width=0.25\linewidth]{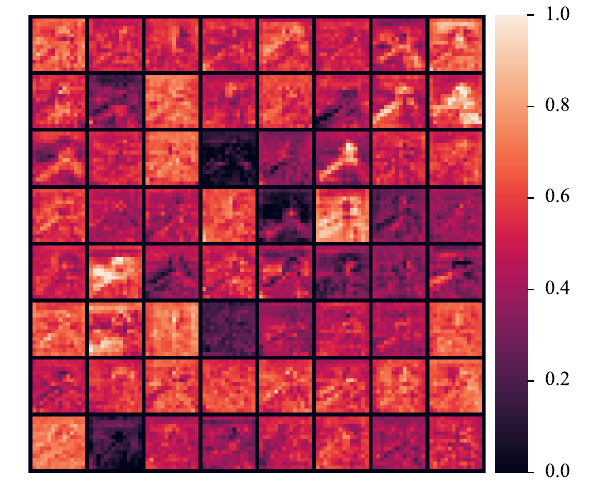}}%
    & \makecell{\includegraphics[width=0.25\linewidth]{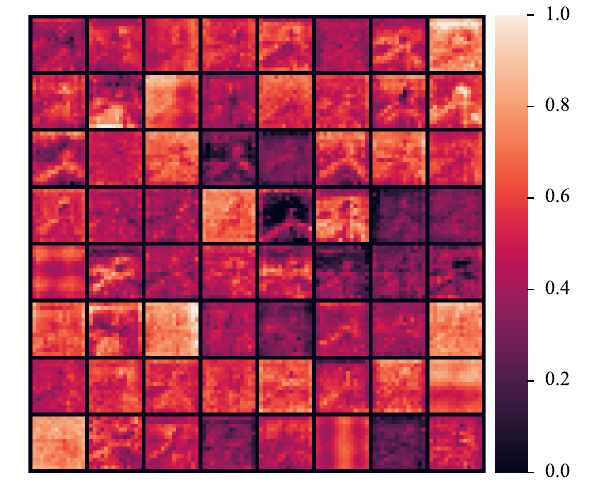}}%
    \\ \midrule

    \makecell{4}
    & \makecell{\includegraphics[width=0.25\linewidth]{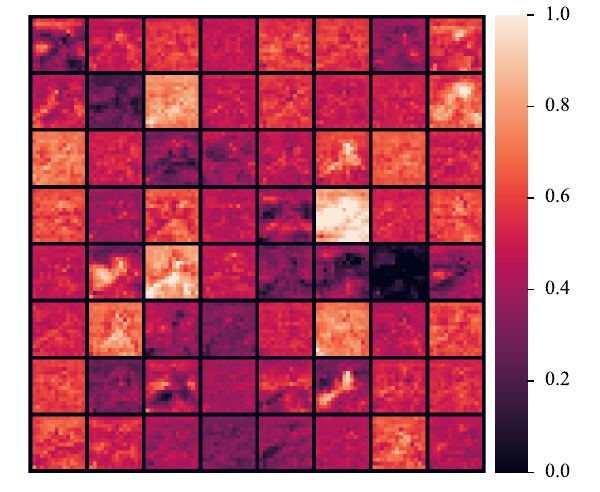}}%
    & \makecell{\includegraphics[width=0.25\linewidth]{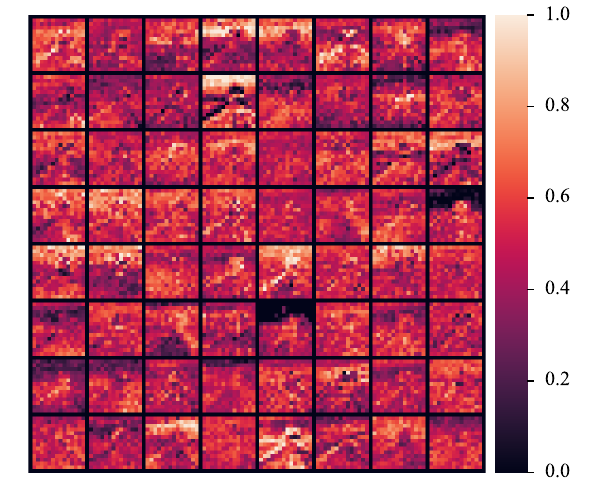}}%
    & \makecell{\includegraphics[width=0.25\linewidth]{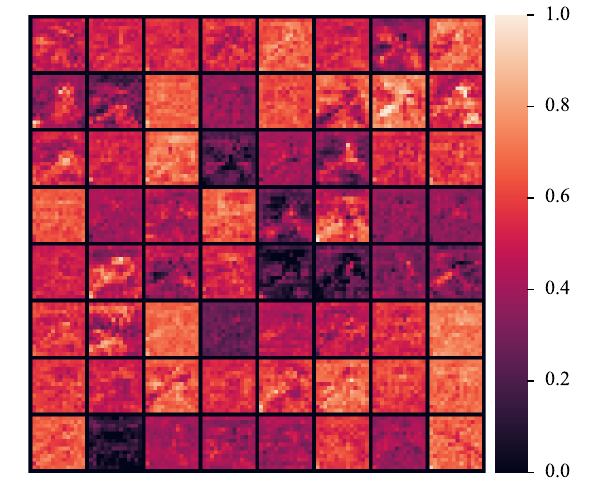}}%
    & \makecell{\includegraphics[width=0.25\linewidth]{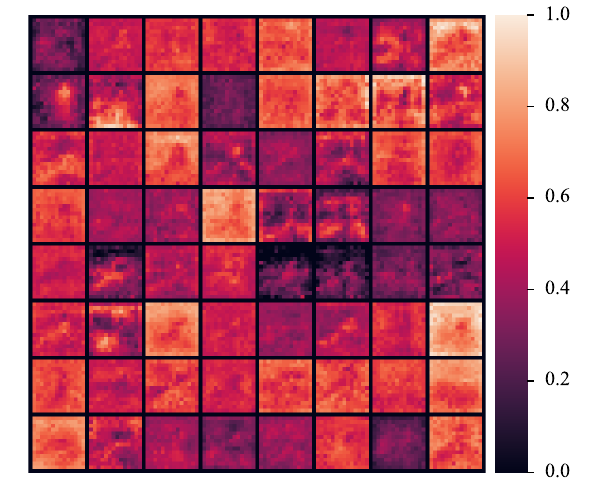}}%
    \\ \midrule

    \makecell{6}
    & \makecell{\includegraphics[width=0.25\linewidth]{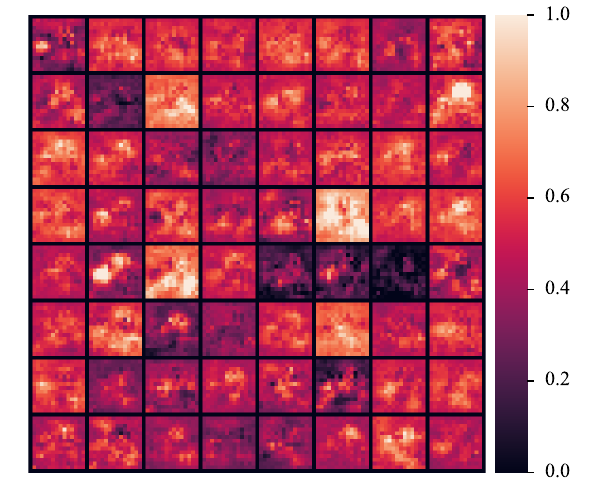}}%
    & \makecell{\includegraphics[width=0.25\linewidth]{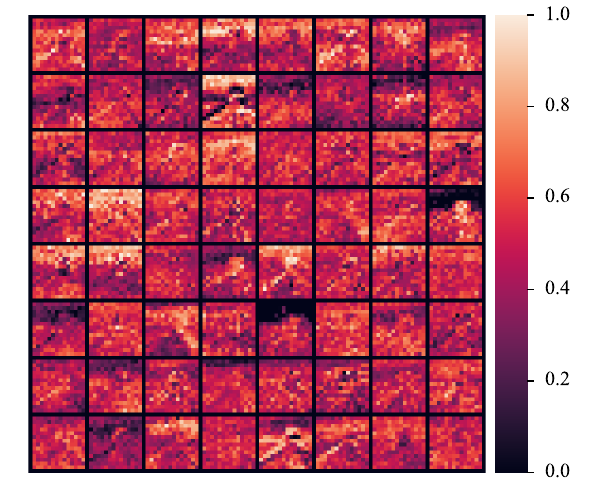}}%
    & \makecell{\includegraphics[width=0.25\linewidth]{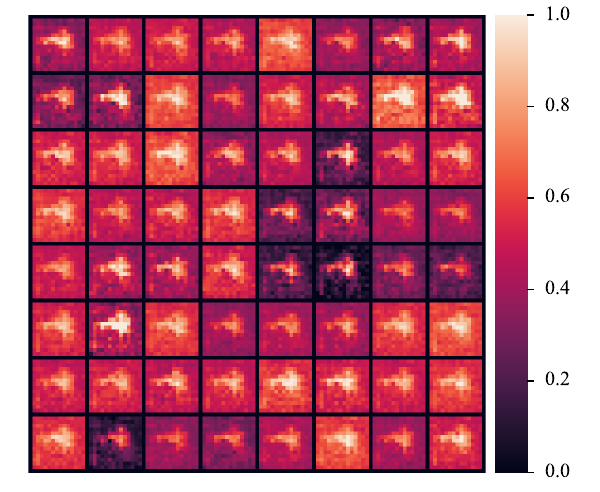}}%
    & \makecell{\includegraphics[width=0.25\linewidth]{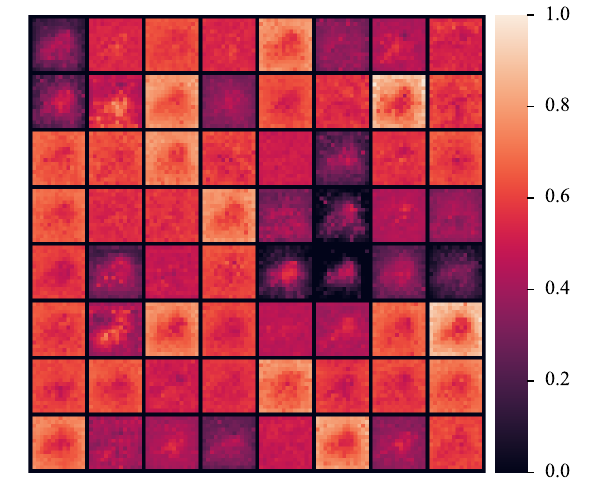}}%
    \\ \bottomrule
\end{tabular}%
}
\label{fig:vis-feat-1}%
\end{table*}

\begin{table*}[!tp]
\centering
\setlength{\tabcolsep}{1pt}%
\caption{Visualization of intermediate features for input~\cref{fig:vis-img-2}. Please zoom for better view.}%
\resizebox{0.9\linewidth}{!}{
\begin{tabular}{l| cc|cc}%
    \toprule
    \textbf{Layer} & \textbf{Agent-S} & \textbf{ViT-S} & \textbf{Agent-S-Joint} & \textbf{ViT-S-Joint} \\ \midrule
    \makecell{2}
    & \makecell{\includegraphics[width=0.25\linewidth]{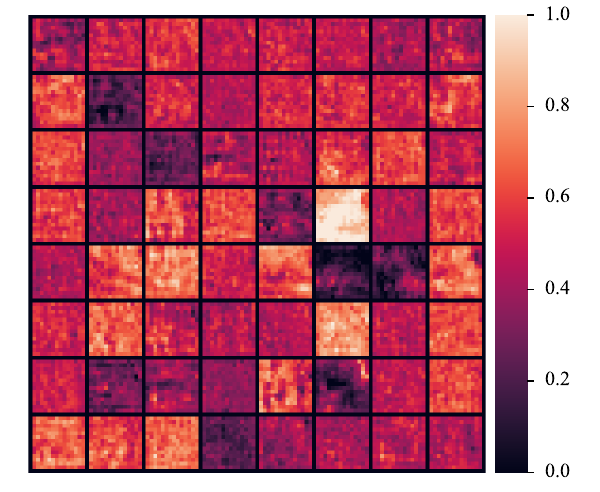}}%
    & \makecell{\includegraphics[width=0.25\linewidth]{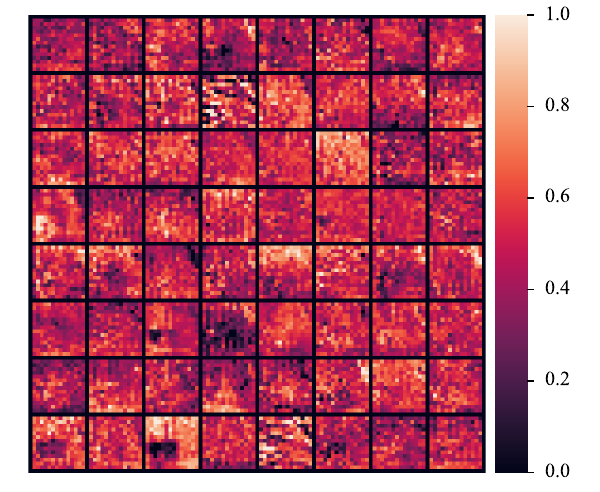}}%
    & \makecell{\includegraphics[width=0.25\linewidth]{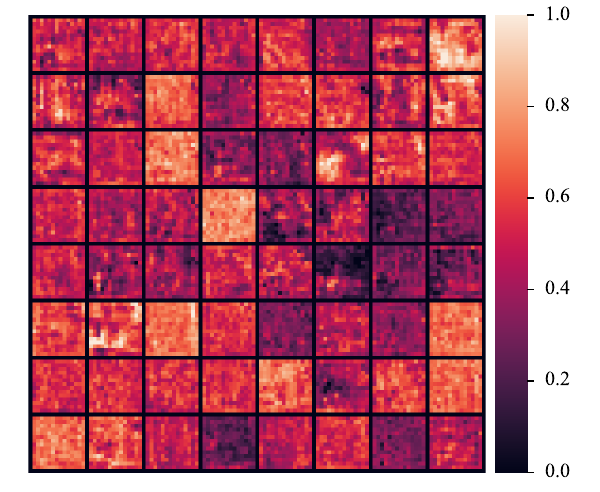}}%
    & \makecell{\includegraphics[width=0.25\linewidth]{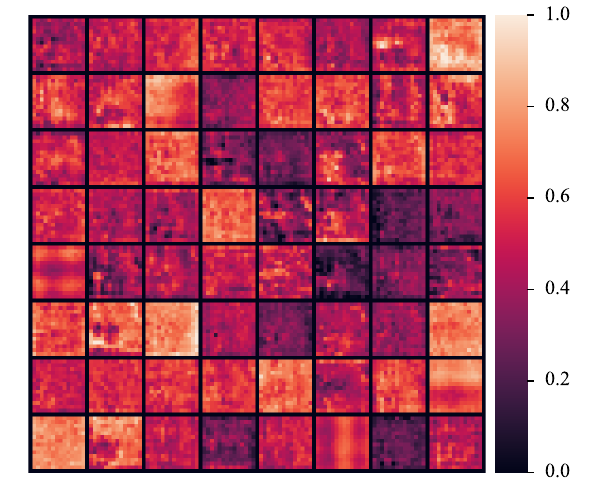}}%
    \\ \midrule

    \makecell{4}
    & \makecell{\includegraphics[width=0.25\linewidth]{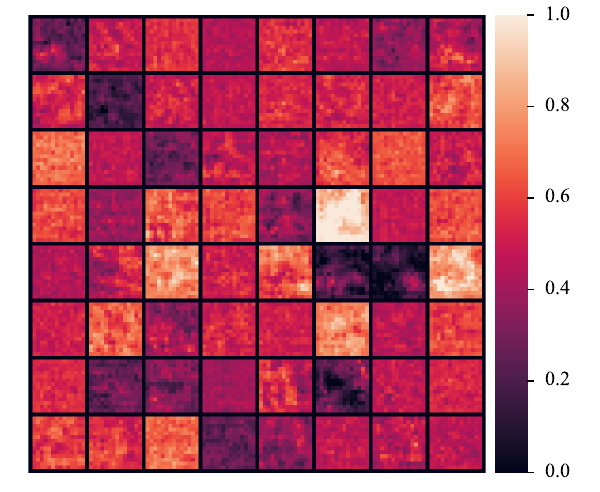}}%
    & \makecell{\includegraphics[width=0.25\linewidth]{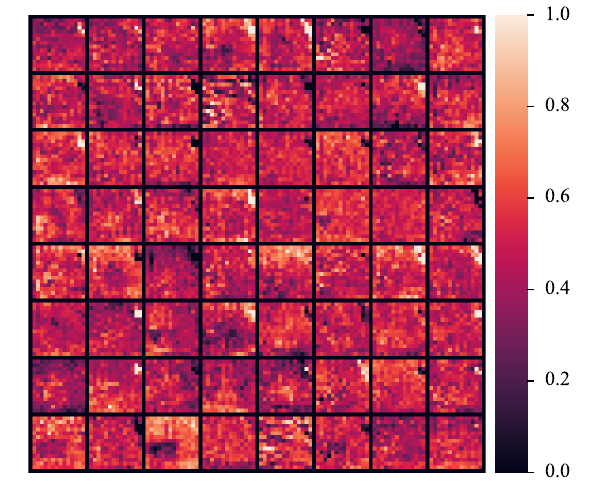}}%
    & \makecell{\includegraphics[width=0.25\linewidth]{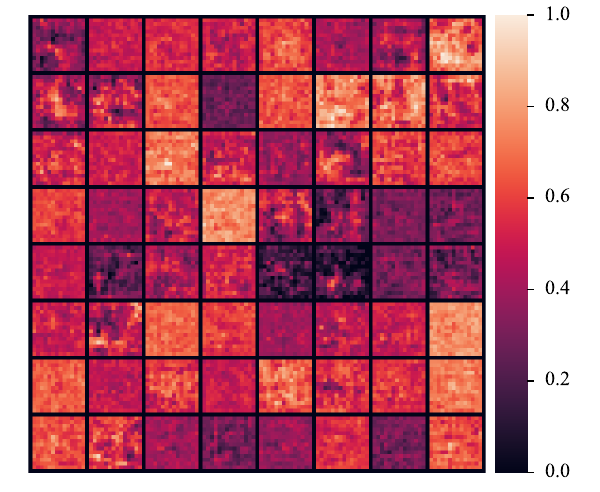}}%
    & \makecell{\includegraphics[width=0.25\linewidth]{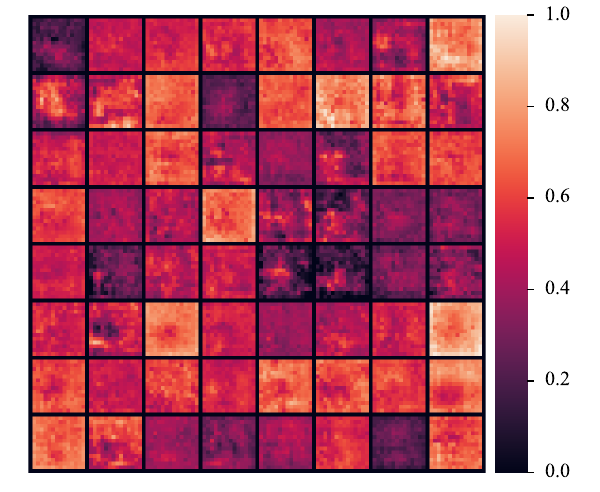}}%
    \\ \midrule

    \makecell{6}
    & \makecell{\includegraphics[width=0.25\linewidth]{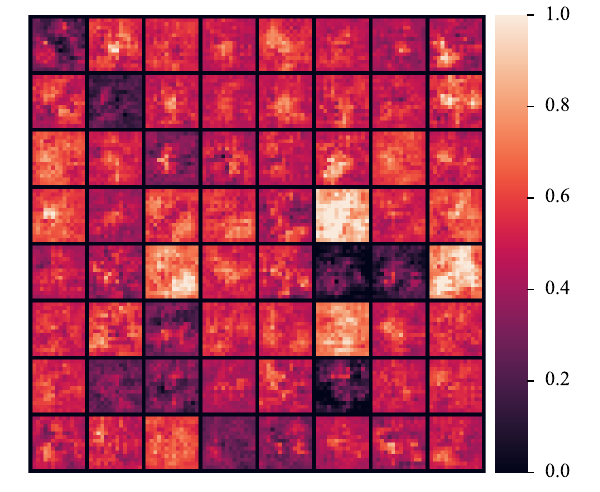}}%
    & \makecell{\includegraphics[width=0.25\linewidth]{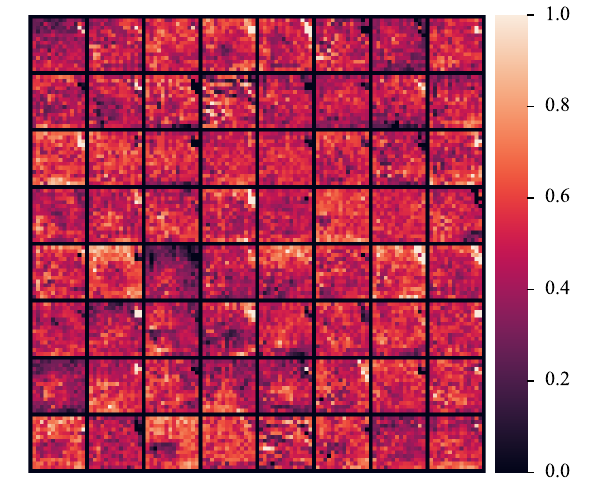}}%
    & \makecell{\includegraphics[width=0.25\linewidth]{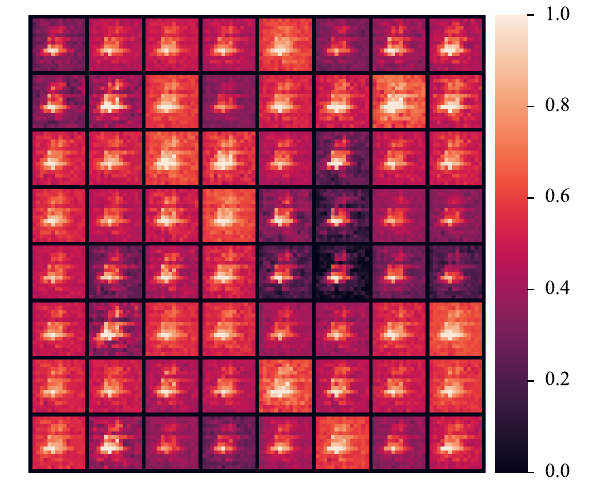}}%
    & \makecell{\includegraphics[width=0.25\linewidth]{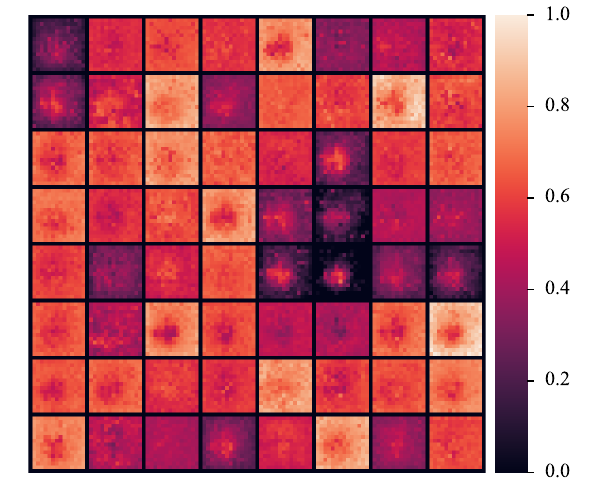}}%
    \\ \bottomrule
\end{tabular}%
}
\label{fig:vis-feat-2}%
\end{table*}